\documentclass{article}
\usepackage[preprint,nonatbib]{neurips23}

\usepackage[utf8]{inputenc} 
\usepackage[T1]{fontenc}    
\usepackage{hyperref}       
\usepackage{url}            
\usepackage{booktabs}       
\usepackage{amsfonts}       
\usepackage{nicefrac}       
\usepackage{microtype}      
\usepackage[table,x11names]{xcolor}
\usepackage{subcaption}
\usepackage{graphicx}
\usepackage[export]{adjustbox}
\usepackage{amsmath}
\usepackage{cleveref}
\usepackage{wrapfig}
\usepackage{enumitem}
\usepackage{multirow}
\usepackage[numbers]{natbib}

\def\our{CASSLE}
\def\guidingnet{Augmentation encoder}
\def\loss{\mathcal{L}}

\title{Augmentation-aware Self-supervised Learning with Conditioned Projector}

\author{%
Marcin Przewięźlikowski$^{1,2,3}$ \thanks{Corresponding author: \texttt{marcin.przewiezlikowski@doctoral.uj.edu.pl}}
\quad Mateusz Pyla$^{1,2,3}$
\quad Bartosz Zieliński$^{1,3}$ \\
\quad \textbf{Bartłomiej Twardowski}$^{3,4,5}$
\quad \textbf{Jacek Tabor}$^1$ 
\quad \textbf{Marek Śmieja}$^1$
\\
$^1$ Jagiellonian University, Faculty of Mathematics and Computer Science\\
$^2$ Jagiellonian University, Doctoral School of Exact and Natural Sciences\\
$^3$ IDEAS NCBR\quad
$^4$ Department of Computer Science, Universitat Autònoma de Barcelona \\
$^5$ Computer Vision Center, Barcelona
}

\begin{document}
\maketitle

\vspace{-0.5cm}
\begin{abstract}
\vspace{-0.25cm}
Self-supervised learning (SSL) is a powerful technique for learning 
from unlabeled data. By learning to remain invariant to applied data augmentations, methods such as SimCLR and MoCo can reach quality on par with supervised approaches. 
{However, this invariance may be detrimental for solving downstream tasks that depend on traits affected by augmentations used during pretraining, such as color.}
In this paper, we propose to foster sensitivity to such characteristics in the representation space by modifying the projector network, a common component of self-supervised architectures. Specifically, we supplement the projector with information about augmentations applied to images. For the projector to take advantage of this auxiliary conditioning when solving the SSL task, the feature extractor learns to preserve the augmentation information in its representations. Our approach, coined \textbf{C}onditional \textbf{A}ugmentation-aware \textbf{S}elf-\textbf{s}upervised \textbf{Le}arning (\our{}), is directly applicable to typical joint-embedding SSL methods regardless of their objective functions. Moreover, it does not require major changes in the network architecture or prior knowledge of downstream tasks. In addition to an analysis of sensitivity towards different data augmentations, we conduct a series of experiments, which show that \our{} improves over various SSL methods, reaching state-of-the-art performance in multiple downstream tasks.\footnote{\href{https://sslneurips23.github.io/paper_pdfs/paper_6.pdf}{A short version of this paper} appeared at the \href{https://sslneurips23.github.io}{NeurIPS 2023 Workshop: Self-Supervised Learning - Theory and Practice}. \href{https://www.sciencedirect.com/science/article/pii/S0950705124012061}{The full paper was published (OA) in 
Knowledge-Based Systems.}} 
\end{abstract}








\section{Introduction}

Artificial neural networks have proven to be a successful family of models in several domains, including, but not limited to, computer vision~\cite{he2015deep}, natural language processing~\cite{gpt}, solving problems at the human level with reinforcement learning~\cite{mnih2015human}, {and biosignal processing in medicine~\cite{kumar2022biosignals}}. This success is attributed largely to their ability to learn useful feature representations~\cite{goodfellow2016deep} without additional effort for input signals preparation. However, training large deep learning models requires extensive amounts of data, which can be costly to prepare, especially when human annotation is needed~\cite{bai2021selfsupervised,kim2022did}.

High-quality image representations can be acquired without relying on explicitly labeled data by
utilizing Self-supervised learning (SSL). A Self-supervised model is trained once on a large dataset without labels and then transferred to different downstream tasks. Initially, self-supervised methods addressed well-defined pretext tasks, such as predicting rotation~\cite{gidaris2018unsupervised} or determining patch position~\cite{doersch2015unsupervised}.
Recent studies in SSL proposed contrastive methods of learning representations that remain invariant when subjected to various data augmentations~\cite{he2020momentum,chen2020simple,chen2021exploring} leading to impressive results that have greatly diminished the disparity with representations learned in a supervised way~\cite{caron2021emerging}.

\begin{figure}[ht]
    \centering
    \includegraphics[width=\textwidth]{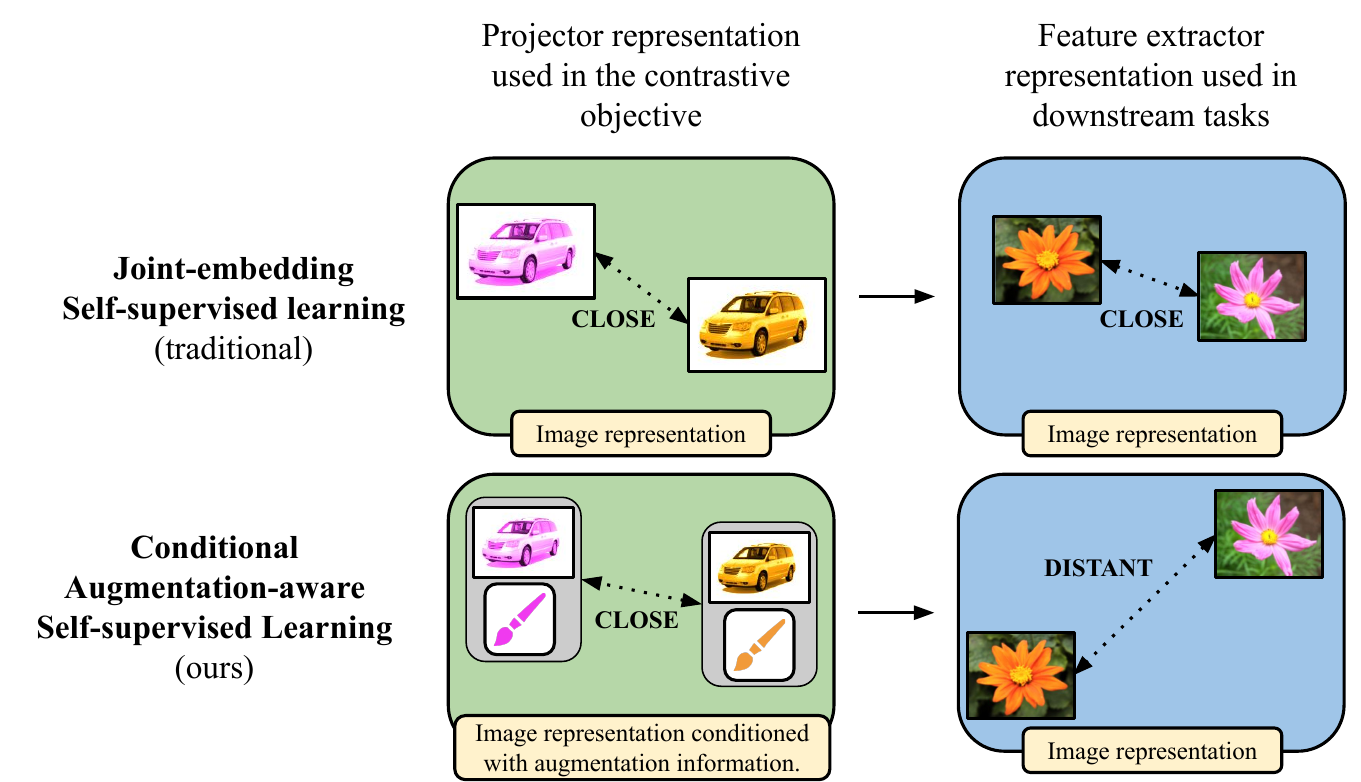}
    \caption{In the traditional self-supervised setting, contrastive loss minimization pulls the representations of augmented image views closer in the latent space of the projector (left). This may also reduce the distance between their feature extractor representations (right). Thus, the representation becomes invariant to augmentation-induced perturbations, which may hinder the performance on downstream tasks. In contrast, the self-supervised objective of \our{} draws together joint representations of images and their augmentations in the projector space (bottom row). By conditioning the projector with augmentation information, image representations retain more sensitivity to perturbations in the feature extractor space. This proves to be beneficial when solving downstream tasks.}
    \label{fig:latent_spaces}
\end{figure}

Nevertheless, contrastive methods 
may perform poorly when a particular downstream task relies on features affected by augmentation~\cite{xiao2020whatshouldnotbecontrastive}. For example, color jittering can result in a representation space invariant to color shifts, which would be detrimental to the task of flower classification (see Figure~\ref{fig:latent_spaces}). Without prior knowledge of possible downstream tasks, this effect is hard to mitigate in contrastive learning~\cite{tian2020makes,xiao2020whatshouldnotbecontrastive}. 
Solutions for retaining information about used data augmentations in the feature extractor representation include forcing it explicitly with a modified training scheme~\cite{xiao2020whatshouldnotbecontrastive,lee2021improving,xie2022whatshouldbeequivariant}, or by preparing a feature extractor to be adapted to a specific downstream task, e.g., with hypernetworks~\cite{chavhan2023amortised}. However, these approaches often involve significant modifications either to the contrastive model architecture~\cite{xiao2020whatshouldnotbecontrastive}, training procedure~\cite{lee2021improving,xie2022whatshouldbeequivariant}, or costly training of additional models~\cite{chavhan2023amortised}.

In this work, we propose a new method called \textbf{C}onditional \textbf{A}ugmentation-aware \textbf{S}elf-\textbf{s}upervised \textbf{Le}arning (\our{}) that mitigates augmentation invariance of representation without neither major changes in network architecture or modifications to the self-supervised training objective. 
We propose to use the augmentation information during the SSL training as additional 
conditioning for the projector network. This encourages the feature extractor network to retain information about augmented image features in its representation. \our{} can be applied to any joint-embedding SSL method regardless of its objective~\cite{chen2020improved,chen2020simple,chen2021exploring,zbontar2021barlow,chen2021empirical}. 
 {The outcome is a general-purpose, augmentation-aware encoder that can be directly used for any downstream task. \our{} presents improved results in comparison to other augmentation-aware SSL methods, improving transferability to downstream tasks where invariance of the model representation to specific data changes could prove detrimental.}

\newpage
\textbf{The main contributions of our work are threefold:
}
\begin{itemize}[leftmargin=*]
    \item We propose a simple yet effective method for Conditional Augmentation-aware Self-supervised Learning (\our{}). Using our 
    conditioned
    projector enables preserving more information about augmentations in representations than in existing methods.
    \item \our{} is a general modification that can be directly applied to existing joint-embedding SSL approaches without introducing additional objectives and major changes in the network architecture.
    \item In a series of experiments we demonstrate that \our{} reaches state-of-the-art performance with different SSL methods for robust representation learning and improves upon the performance of previous augmentation-aware approaches. Furthermore, our analysis indicates that \our{} learns representations with increased augmentation sensitivity compared to other approaches.
\end{itemize}

{This manuscript is structured as follows: In Section~\ref{sec:related}, we give an overview of prior works in SSL, with a special focus on the line of work on augmentation-aware SSL. Section~\ref{sec:presentation_of_our} presents our proposed technique, \our{}. In Section~\ref{sec:experiments}, we conduct a thorough experimental analysis of \our{}, in terms of its performance and unique properties. Section~\ref{sec:conclusion} concludes the manuscript.}

\section{Related work}
\label{sec:related}

{In this section, we describe prior works on self-supervised learning, with a focus on computer vision. We then recall the line of work on augmentation-awareness of self-supervised models, which motivates our paper. Finally, we briefly compare our proposed method with techniques proposed in previous works.}

\paragraph{Self-supervised learning} (SSL) is a paradigm of learning representations from unlabeled data that can later be used for downstream tasks defined by human annotations \cite{saleh2022selfsupervisedsurvey,balestriero2023cookbook}. 
Despite learning artificial \emph{pretext tasks}, instead of data-defined ones, SSL models have achieved tremendous success in a plethora of domains \cite{devlin2019bert,wickstrom2022mixing,schiappa2022self,Bengar2021ICCV}. This includes computer vision, where
a variety of pretext tasks has been proposed \cite{doersch2015unsupervised,zhang2016colorful,noroozi2017unsupervised,gidaris2018unsupervised}.
{However, arguably the most prominent and successful SSL technique to emerge in recent years is the training of joint-embedding models for augmentation invariance \cite{becker1992self,oord2019representation}, defined by objectives such as contrastive InfoNCE (Information Noise-Contrastive Estimation) loss \cite{he2020momentum,chen2020simple,chen2020improved}, self-distillation  \cite{grill2020bootstrap,chen2020simple,oquab2023dinov2} or CCA (Canonical Correlation Analysis) \cite{caron2020unsupervised,zbontar2021barlow,bardes2022vicreg}.} 
Those objectives are often collectively referred to as \emph{contrastive objectives} \cite{tian2022understanding,balestriero2023cookbook}. 
A common component of joint-embedding architectures is the \emph{projector network}, which maps representations of the feature extractor into the space where the contrastive objective is imposed \cite{chen2020simple,chen2020improved}. The usefulness of the projector has been explained through the lens of transfer learning, where it is often better to transfer intermediate network representations, to reduce the biases from the pretraining task \cite{yosinski2014transferable,bordes2022guillotine}. The projector also helps to mitigate the noisy data augmentations and enforces some degree of pairwise independence of image features \cite{balestriero2023cookbook,mialon2022variance}.

\paragraph{Augmentation invariance of self-supervised models} is a natural consequence of training them with contrastive objectives, as SSL methods are prone to suppressing features that are not useful for optimizing the contrastive objectives~\cite{chen2021intriguing,robinson2021contrastive}. 
While a common set of augmentations demonstrated to typically work well on natural images in SSL has been established in the literature~\cite{he2020momentum,chen2020simple,chen2021exploring,caron2020unsupervised,zini2023planckian}, the optimal choice of augmentations varies between specific tasks~\cite{tian2020makes,ericsson2022selfsupervised}.
\cite{xiao2020whatshouldnotbecontrastive} find that augmentation invariance can hinder the {model performance} on downstream tasks that require attention to precisely those traits that it had been previously trained to be invariant to. On the other hand, \cite{zhang2019aet} observe that the objective of predicting augmentation parameters {can in itself be} a useful pretext task for SSL.
Those works inspired several techniques of retaining augmentation-specific information in joint-embedding models, such as projectors sensitive to different augmentation types~\cite{xiao2020whatshouldnotbecontrastive,ericsson2022selfsupervised}, adding an objective of explicit prediction of augmentation parameters~\cite{lee2021improving}, 
, as well as task-specific pretraining~\cite{raghu2021metalearning,wagner2022importance}. 
The above approaches produce general-purpose feature extractors that can be transferred to downstream tasks without further tuning their parameters. However, they often involve complex modifications either to the SSL model architecture~\cite{xiao2020whatshouldnotbecontrastive}, training procedure~\cite{lee2021improving,xie2022whatshouldbeequivariant}, or simply tedious task-specific pretraining~\cite{wagner2022importance}. Another line of work proposes to train Hypernetworks~\cite{ha2016hypernetworks} which produce feature extractors invariant to chosen subsets of augmentations -- a more elastic, but considerably harder to train approach \cite{chavhan2023amortised}. 
Several works have proposed fostering the equivariance of representations to data transformations using augmentation information.{\cite{xie2022whatshouldbeequivariant} modulate the contrastive objective with augmentation strength, \cite{bhardwaj2023steerable} use augmentation information as a signal for equivariance regularization in the supervised setting, whereas \cite{garrido2023self} extend the VicReg (Variance-Invariance-Covariance Regularization)~\cite{bardes2022vicreg} objective with a predictor whose parameters are generated from augmentation information by a hypernetwork~\cite{ha2016hypernetworks}.}

{
\paragraph{Novelty of \our{}}
We follow the line of works that aim to use augmentation information to improve the Self-supervised training of general-purpose natural image feature extractors~\cite{xiao2020whatshouldnotbecontrastive,lee2021improving,chavhan2023amortised}.
Contrary to the above works, we do not make any modifications to the objective functions of the extended approaches. This removes the need to balance the invariance and sensitivity objectives, present in~\cite{lee2021improving,garrido2023self}. As opposed to~\cite{xiao2020whatshouldnotbecontrastive}, we optimize only a single invariance objective, instead of multiple objectives dedicated to each augmentation type. Compared to~\cite{chavhan2023amortised}, our method of conditioning is more straightforward to integrate into SSL training mechanisms, as we do not require all samples in a data batch to be augmented in the same way. We also do not make any modifications to the architecture of the feature extractor, unlike~\cite{chavhan2023amortised}, where the feature extractor parameters are set based on augmentation parameters. In \our{}, the feature extractor is thus much easier to use out of the box.  

From a technical perspective,
we compare \our{} and other augmentation-aware SSL approaches on a wide variety of image processing tasks, including image classification, object detection, image retrieval, and rotation prediction. Moreover, we show how the InfoNCE can serve as a useful measure for comparing augmentation-awareness.
To the best of our knowledge, we are the first to identify the difference between augmented and non-augmented data as a useful conditioning signal.  We are also the first to analyze the masked image modeling of SSL models in terms of augmentation-awareness.
Overall, our experiments provide a comparison of a variety of augmentation-aware approaches, providing guidelines for SSL practitioners.

 }

\section{Method}
\label{sec:presentation_of_our}

In this section, we present our approach,  \textbf{C}onditional \textbf{A}ugmentation-aware \textbf{S}elf-\textbf{s}upervised \textbf{le}arning (\our{}). Section \ref{sec:premiminaries} provides background on joint-embedding self-supervised methods and their limitations. Section \ref{sec:our} explains the essence of \our{} and how it leverages augmentation information to improve the quality of learned representations. 
Section \ref{sec:guide_hyperparams} details the practical implementation of \our{}'s conditioning mechanism.

\begin{table}[h]
    \centering
    
    \caption{{Glossary of the key concepts used in this manuscript.}}
      \label{tab:reference}

      \label{tab:notation}
    \resizebox{\linewidth}{!}{
      \begin{tabular}{ccl} %
        \toprule
       & \textbf{Symbol} & \multicolumn{1}{c}{\textbf{Explanation}} \\

        \midrule
         \multirow{4}{*}{\rotatebox{90}{\textbf{Variables}}} & $\mathbf{x} \in \mathcal{X} \subset \mathbb{R}^D$ &  Data sample \\ 
          & $\omega \in \Omega \subset \mathbb{R}^A$ &  Augmentation information  vector \\
          & $\mathbf{v} \in \mathcal{X} \subset \mathbb{R}^D$ &  Augmented data sample \\
         & $\mathbf{e} \in \mathbb{R}^E$ & Feature extractor embedding \\
        \midrule
          \multirow{5}{*}{\rotatebox{90}{\textbf{Functions}}} & $t_\omega: \mathcal{X} \rightarrow \mathcal{X}$ & Augmentation parametrized by $\omega$ \\
         & $f: \mathcal{X} \rightarrow \mathbb{R}^E $ & Feature extractor \\
        & $\gamma: \Omega \rightarrow \mathbb{R}^G$ & \guidingnet{} \\
        &  $\pi: \mathbb{R}^{E+G} \rightarrow \mathbb{R}^k $ & Projector \\
        & $\loss :  \mathbb{R}^k \times \mathbb{R}^k \rightarrow \mathbb{R}$ & Objective function in joint-embedding learning \\
         \midrule
         \multirow{9}{*}{\rotatebox{90}{\textbf{Dimensions}}} & $D$ &  Image shape, i.e. $D=224 \times 224 \times 3$ \\
        & \multirow{2}{*}{$A$} &  Augmentation vector size -- depends on the choice of   \\ 
        & &  augmentations; in our case, $A=14$ (see Section~\ref{sec:guide_hyperparams})\\
        & \multirow{2}{*}{$E$} & Feature extractor embedding size  -- depends on the architecture;\\  
        & &  e.g. for Resnet-50~\cite{he2015deep}, $E=2048$\\
        & \multirow{2}{*}{$G$} & \guidingnet{} embedding size -- hyperparameter; \\
        & & in our case, $G=64$ \\
        &  \multirow{2}{*}{$k$} & Projector embedding size -- depends on the SSL method; \\ 
        & & e.g. for MoCo-v2, $k=128$ (see Table~\ref{tab:hyperparams})  \\

        \bottomrule
      \end{tabular}
      }
  
\end{table}

\subsection{Preliminaries} 
\label{sec:premiminaries}

{In this section, we introduce the key concepts in joint-embedding SSL. We include Table~\ref{tab:reference} as a reference of key concepts used in this manuscript.}
A typical 
joint-embedding
framework used in self-supervised learning consists of an augmentation function $t_\omega$ and two networks: feature extractor $f$ and projector $\pi$.
Let $\mathbf{v}_1 = t_{\mathbf{\omega}_1}(\mathbf{x}), \mathbf{v}_2 = t_{\mathbf{\omega}_2}(\mathbf{x})$ be two augmentations of a sample $\mathbf{x} \sim \mathcal{X}$ parameterized by $\mathbf{\omega}_1, \mathbf{\omega}_2 \sim \Omega$. 
{The feature extractor maps them into the embedding space, i.e. the representation used in downstream tasks.} To make the representation invariant to data augmentations, $\mathbf{e}_1 = f(\mathbf{v}_1)$ is forced to be similar to $\mathbf{e}_2 = f(\mathbf{v}_2)$. {This can be expressed by various objective functions~\cite{garrido2022duality}, such as contrastive InfoNCE~\cite{oord2019representation,chen2020simple,he2020momentum}\footnote{
While contrastive objectives such as InfoNCE~\cite{oord2019representation} regularize the representation using negative pairs, we omit them from our notation for the sake of brevity.}, or CCA~\cite{zbontar2021barlow} -- which we collectively denote as $\loss$.}
{Instead of imposing the objective $\loss$ directly on the embedding space of $f$, a projector $\pi$ transforms the embeddings into another representation space, where $\loss$ is applied.}
This trick, known as \emph{Guillotine Regularization}, helps the feature extractor to better generalize to downstream tasks, due to $f$ not being directly affected by $\loss$~\cite{yosinski2014transferable,chen2020simple,chen2020improved,bordes2022guillotine}.

Minimizing $\loss(\pi(\mathbf{e}_1), \pi(\mathbf{e}_2))$ directly leads to reducing the distance between embeddings $\pi(\mathbf{e}_1)$ and $\pi(\mathbf{e}_2)$. However, $\loss$ still indirectly encourages the intermediate network representations (including the output of the feature extractor $f$) to also conform to the objective to some extent. As a result, the feature extractor tends to erase the information about augmentation from its output representation. This behavior may however be detrimental for certain downstream tasks (see Figures \ref{fig:latent_spaces} and \ref{fig:exp_invariance}), which rely on features affected by augmentations. For instance, learning invariance to color jittering through standard contrastive methods may lead to degraded performance on the downstream task of flower recognition, which is not a color-invariant task~\cite{tian2020makes,xiao2020whatshouldnotbecontrastive}. Thus, the success of typical SSL approaches depends critically on a careful choice of augmentations used for model pretraining \cite{chen2020simple,tian2020makes}.


\subsection{\our{}} 
\label{sec:our}

\begin{figure*}[t]
    \centering
    \includegraphics[width=\textwidth]{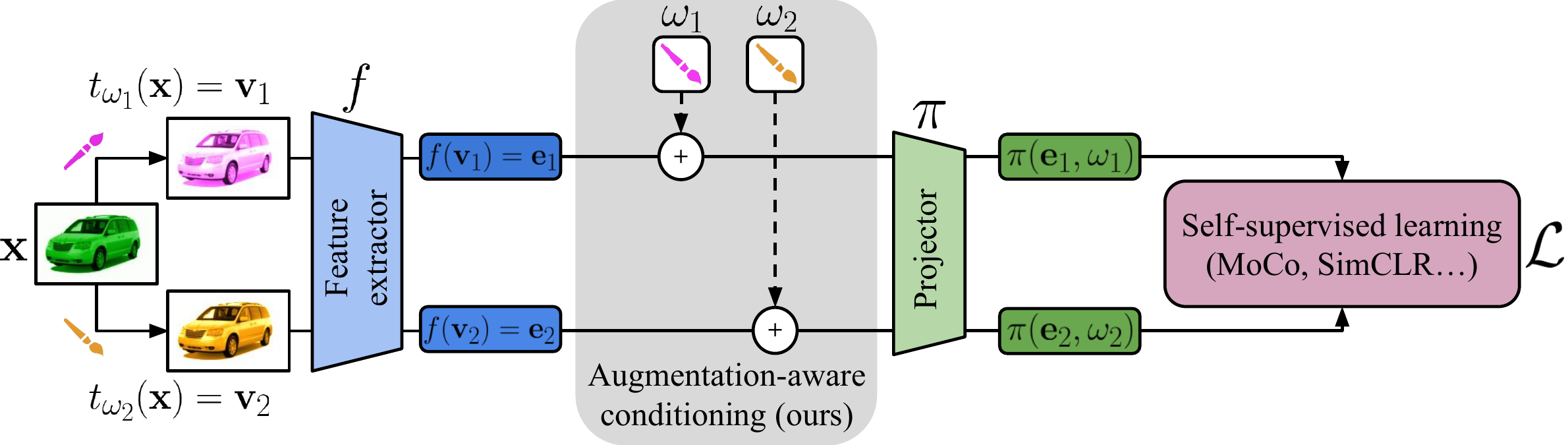}
    \caption{Overview of \our{}. We extend the typical self-supervised learning approaches by incorporating the information of augmentations applied to images into the projector network. 
    In \our{}, the SSL objective is thus imposed on joint representations of images and the augmentations that had been applied to them. 
    This way, \our{} enables the feature extractor to be more aware of augmentations than the methods that do not condition the projector network.}   
    \label{fig:architecture}
    \vspace{-0.5cm}
\end{figure*}

{We have reviewed Self-supervised learning (SSL) and the limitations of joint-embedding approaches.} To overcome the above limitations, we facilitate the feature extractor to encode the information about augmentations in its output representation. In consequence, the obtained representation will be more informative for downstream tasks that depend on features modified by augmentations.

\our{} achieves this goal by conditioning the projector $\pi$ on the parameters of augmentations used to perturb the input image. 
Specifically, we modify $\pi$ so that apart from embedding $\mathbf{e}$, it also receives augmentation information $\omega$ and projects their joint representation into the space where the objective $\loss$ is imposed.
We do not alter the $\loss$ itself; instead, training relies on minimizing the contrastive loss $\loss$ between $\pi(\mathbf{e}_1 | \mathbf{\omega}_1)$ and $\pi(\mathbf{e}_2 | \mathbf{\omega}_2)$. Thus, $\pi$ learns to draw $\mathbf{e}_1$ and $\mathbf{e}_2$ together in its representation space \emph{on condition of} $\mathbf{\omega}_1$ \emph{and} $\mathbf{\omega}_2$. 
We visualize the architecture of \our{} in Figure~\ref{fig:architecture}.

\newpage
We provide a rationale for why \our{} preserves information about augmented features in the representation space. Since augmentation information vectors $\omega$ do not carry any information about source images $\mathbf{x}$, their usefulness during pretraining could be explained only by using knowledge of transformations $t_\omega$ that had been applied to $\mathbf{x}$ to form views $\mathbf{v}$. However, for such knowledge to be acted upon, features affected by $t_\omega$ must be preserved in the feature extractor representation $f(\mathbf{v})$.

Let us assume the opposite, that $\omega$ is not useful for \our{} to solve the task defined by $\loss$. If this were the case, then for any $\omega_3 \sim \Omega$ the following would hold:
\begin{equation}
    \label{eq:probs_are_equal}
    p(\pi(\mathbf{e}_1|\omega_1) | \pi(\mathbf{e}_2|\omega_2)) = p(\pi(\mathbf{e}_1|\omega_1) | \pi(\mathbf{e}_2|\omega_3)).
\end{equation}

$p(\pi(\mathbf{e}_1|\omega_1) | \pi(\mathbf{e}_2|\omega_2))$ can be understood as conditional probability that \emph{$\pi(\mathbf{e}_1|\omega_1)$ is a representation of an image $\mathbf{x}$ transformed by $t_{\omega_1}$, given the knowledge that $\pi(\mathbf{e}_2|\omega_2)$ is a representation of $\mathbf{x}$ transformed by $t_{\omega_2}$}. Equation \ref{eq:probs_are_equal} implies that replacing the knowledge of $t_{\omega_2}$ with any other randomly sampled $t_{\omega_3}$ does not affect the inference process of \our{}. 

\begin{figure}[h]
         \centering
         \hfill
          \hspace{0.6cm}
         \includegraphics[width=0.45\textwidth]{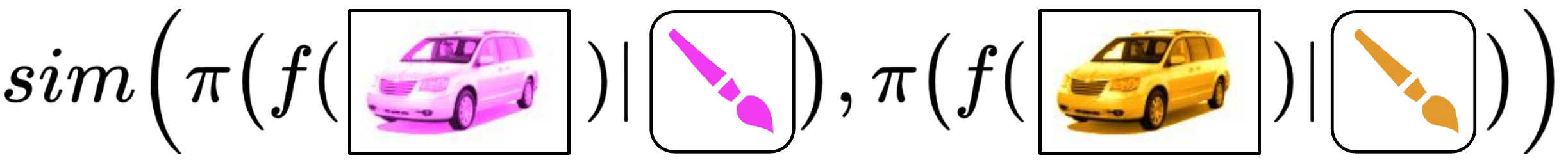} 
         \hspace{0.4cm}
         \includegraphics[width=0.45\textwidth]{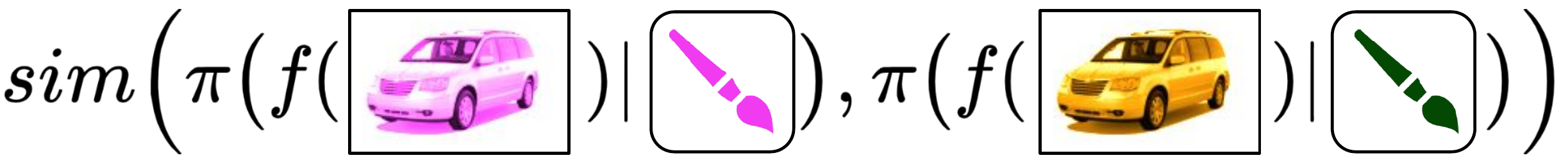} 
         \\
        \includegraphics[width=0.6\textwidth]{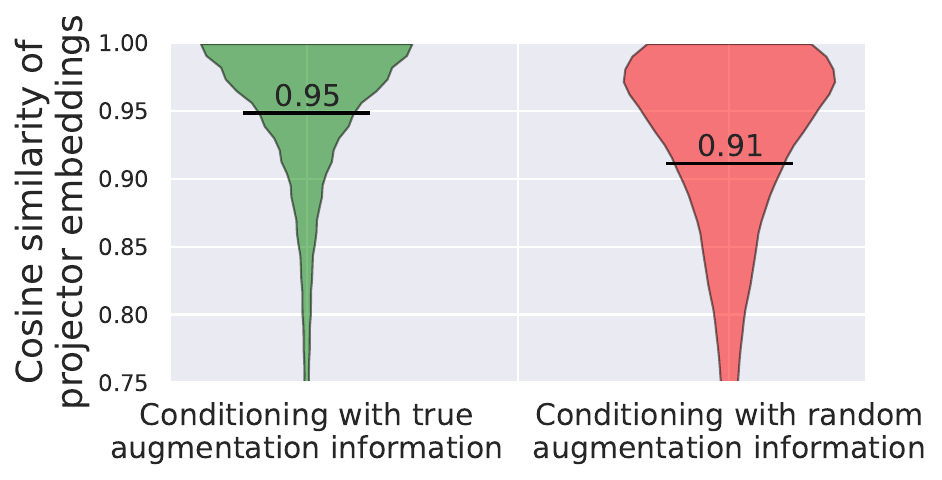}
        \caption{
        {Cosine similarities of \our{} projector $(\pi)$ representations when conditioned with augmentation information from either their respective images {\color{Green4}(green)} or randomly sampled {\color{red}(red)}.} Solid lines denote the mean values of similarities. 
        Conditioning the \our{} projector with wrong augmentation information decreases its ability to draw image pairs together, indicating that it indeed relies on augmentation information to perform its task.
        }
    \label{fig:exp:conditioning}    
    \vspace{-0.25cm}
\end{figure}

To demonstrate that this is not the case, we measure in Figure \ref{fig:exp:conditioning} the {cosine similarity 
(denoted as $sim; sim(\mathbf{a},\mathbf{b}) = \frac{\mathbf{a}^\top \mathbf{b}}{\lVert \mathbf{a} \rVert \lVert \mathbf{b}\rVert}$)} 
of projector representations of 5000 positive augmented image pairs from the ImageNet-100 test set, {using a model trained with \our{}}.
{The green plot denotes the similarities of representations conditioned on the parameters of the augmentations used to construct those image pairs. On the other hand, in the red plot, one of the projector representations is conditioned with \emph{false} augmentation parameters, i.e. randomly sampled parameters that are unrelated to the augmented image.}

It is evident from Figure \ref{fig:exp:conditioning} that the cosine similarity of embeddings decreases when false  augmentation parameters ($\omega_3$ instead of $\omega_2$) are supplied to the projector, i.e:
\begin{equation}
\label{eq:similarities_leq}
\mathbb{E}_{x \sim \mathbb{X}, \{\omega_1, \omega_2, \omega_3 \} \sim \Omega  } [ 
sim\left(\pi(\mathbf{e}_1|\omega_1), \pi(\mathbf{e}_2|\omega_2)\right)  
-  
sim \left(\pi(\mathbf{e}_1|\omega_1), \pi(\mathbf{e}_2|\omega_3) \right) 
] > 0.
\end{equation}
{Recall that in contrastive SSL, the cosine similarity of embeddings corresponds to their probability density. This is because the InfoNCE loss is formulated as cross-entropy, where the activation function is defined as cosine similarity between respective image embeddings, and class labels are replaced with the indices of corresponding positive embedding pairs~\cite{oord2019representation,he2020momentum,chen2020simple}.} Hence, minimizing $\loss$ leads to:
\begin{equation}
    \label{eq:sim_is_density}
    sim\left(\pi(\mathbf{e}_1|\omega_1), \pi(\mathbf{e}_2|\omega_2)\right)
    \quad
    \propto
    \quad
    \frac{p(\pi(\mathbf{e}_1|\omega_1) | \pi(\mathbf{e}_2|\omega_2))}{p(\pi(\mathbf{e}_1|\omega_1))}.
\end{equation}
It follows from  \ref{eq:similarities_leq} that, in practice
\begin{equation}
    \label{eq:densities_leq}
    \mathbb{E}_{x \sim \mathbb{X}, \{\omega_1, \omega_2, \omega_3 \} \sim \Omega  } \left[
    \frac{p(\pi(\mathbf{e}_1|\omega_1) | \pi(\mathbf{e}_2|\omega_2)) - p( \pi(\mathbf{e}_1|\omega_1) | \pi(\mathbf{e}_2|\omega_3) )}{p(\pi(\mathbf{e}_1|\omega_1))}
    \right]
    > 0
\end{equation}
and, since $p(\pi(\mathbf{e}_1|\omega_1)) > 0$, 
\begin{equation}
    \mathbb{E}_{x \sim \mathbb{X}, \{\omega_1, \omega_2, \omega_3 \} \sim \Omega  } \left[
    p(\pi(\mathbf{e}_1|\omega_1) | \pi(\mathbf{e}_2|\omega_2))  -
    p(\pi(\mathbf{e}_1|\omega_1) | \pi(\mathbf{e}_2|\omega_3))
    \right] > 0. 
\end{equation}
Moreover, we measure whether $p(\pi(\mathbf{e}_1|\omega_1) | \pi(\mathbf{e}_2|\omega_2)) > p(\pi(\mathbf{e}_1|\omega_1) | \pi(\mathbf{e}_2|\omega_3))$ for each of the considered image view pairs and find it to be true in 92\% of the considered cases.

In \our{}, \emph{the conditional probability of matching a positive pair of image representations increases when the correct augmentation information is known}, {which implies that information describing the augmented features is indeed preserved in the representation of its feature extractor}.

\our{} can be applied to a variety of joint-embedding SSL methods, as the only practical modification it makes is changing the projector network to utilize the additional input $\omega$, describing the augmentations.
We do not modify any other aspects of the self-supervised approaches, such as objective functions, which is appealing from a practical perspective. 
Last but not least, the architecture of the feature extractor in \our{} is not affected by the introduced augmentation conditioning, as we only modify the input to the projector, which is discarded after the pretraining. Just like in vanilla SSL techniques, the feature extractor can be directly re-used for downstream tasks.

\subsection{Practical implementation of the conditioning mechanism}
\label{sec:guide_hyperparams}

{We have introduced \our{} and described the rationale behind this method. In this section, we discuss the practical aspects of the conditioning with augmentation information -- the core component of \our{}. }

In this work, we focus on a set of augmentations used commonly in the literature~\cite{chen2020simple,chen2020improved,chen2021exploring}, listed below along with descriptions of their respective parameters $\omega^{aug}$:
\begin{itemize}[leftmargin=*]
    \item \textbf{random cropping} -- $\mathbf{\omega}^{c} \in [0,1]^4$ describes the normalized coordinates of cropped image center and cropping sizes.
    \item \textbf{color jittering} -- $\mathbf{\omega}^{j} \in [0,1]^4$ describes the normalized intensities of brightness, contrast, saturation, and hue adjustment.
    \item \textbf{Gaussian blurring} -- $\mathbf{\omega}^{b} \in [0,1]$ is the standard deviation of the Gaussian filter used during the blurring operation.
    \item \textbf{random horizontal flipping} -- $\mathbf{\omega}^{f} \in \{0,1\}$ indicates whether the image has been flipped.
    \item \textbf{random grayscaling} -- $\mathbf{\omega}^{g} \in \{0,1\}$ indicates whether the image has been reduced to grayscale.
\end{itemize}
To enhance the projector's awareness of the color changes in the augmented images, we additionally enrich $\mathbf{\omega}$ with information about \textbf{color difference} -- $\mathbf{\omega}^{d} \in [0,1]^3$, which is computed as the difference between the mean values of color channels of the image before and after the color jittering operation. We empirically demonstrate that inclusion of $\mathbf{\omega}^{d}$ in $\mathbf{\omega}$ improves the performance of \our{} (see Section \ref{sec:exp:ablation}).

{We construct augmentation information $\mathbf{\omega} \in \Omega$ by concatenating vectors $\mathbf{\omega}^{aug}$ describing the parameters of each augmentation type~\cite{lee2021improving}. Since each of the above $\omega^{aug}$ can be expressed as a vector of one or more scalars, the concatenated augmentation information vectors contain 14 scalars, i.e. $\omega \in \Omega \subset \mathbb{R}^{14}$. We also explore withholding some augmentation information during training (resulting in an appropriately reduced size of $\omega$), but find that conditioning on full augmentation information leads to the best representation quality (see Section \ref{sec:exp:ablation}).} 

\begin{figure}[h]
    \centering
    \includegraphics[width=\textwidth]{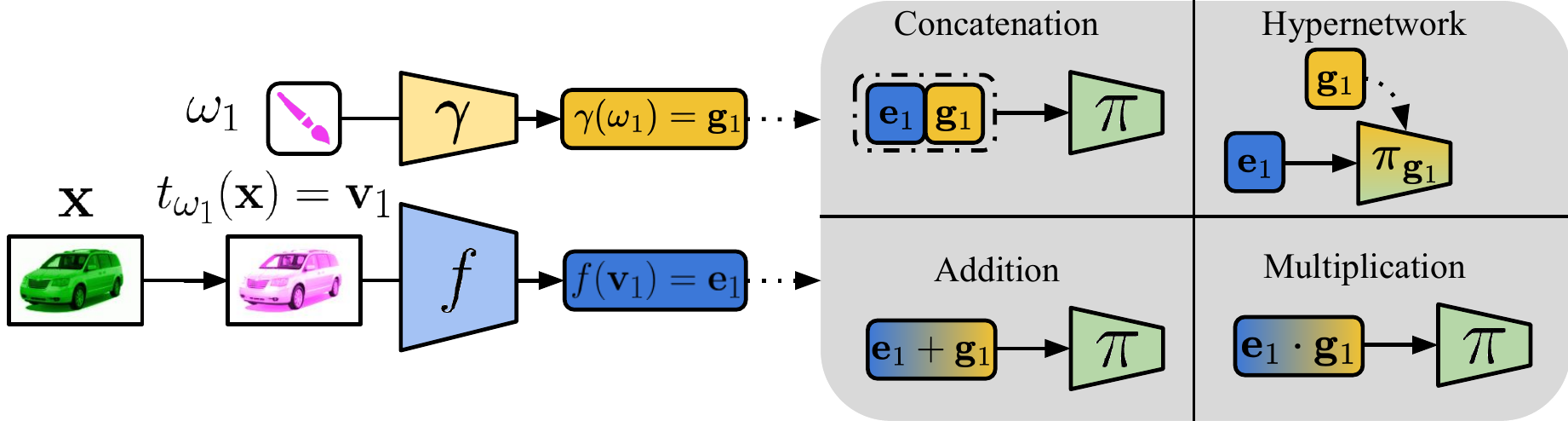}
    \caption{{A visualization of four considered methods of conditioning the projector with augmentation information.}}
    \label{fig:conditioning_details}
\end{figure}

We consider four methods of 
injecting $\omega$ into $\pi$:
joining $\omega$ and $\mathbf{e}$ through \textbf{(i)} concatenation, modulating $\mathbf{e}$ with $\omega$ through element-wise \textbf{(ii)} addition or \textbf{(iii)} multiplication, or \textbf{(iv)} using $\omega$ as an input to a hypernetwork \cite{ha2016hypernetworks} which generates the parameters of $\pi$. Apart from concatenation, all of those methods require transforming $\omega$ into \emph{augmentation embeddings} $\mathbf{g} \in \mathcal{G}$ of shapes required by the
conditioning
operation. For example, when modulating $\mathbf{e}$ with $\mathbf{g}$, dimensions of $\mathbf{e}$ and  $\mathbf{g}$ must be equal.
For this purpose, we precede the projector with additional \emph{\guidingnet{}} $\gamma: \Omega \rightarrow \mathcal{G}$. For the architecture of $\gamma$ we choose the Multilayer Perceptron. An additional advantage of $\gamma$ is that it allows for learning a representation of $\omega$ which is more expressive for processing by $\pi$, which is typically a shallow network. {We summarize the above conditioning mechanisms in Figure~\ref{fig:conditioning_details}}.
In practice, we find that conditioning $\pi$ through the concatenation of $\mathbf{e}$ and $\mathbf{g}$ yields the best-performing representation (see Section \ref{sec:exp:ablation}).

\newpage
\section{Experiments}
\label{sec:experiments}
\vspace{-0.25cm}
{We have introduced \our{} and described the theoretical and practical aspects of its design. In this Section, we conduct an experimental analysis of our approach, which we use to extend several recent SSL frameworks -- MoCo (Momentum Contrastive Learning)~\cite{he2020momentum,chen2020improved,chen2021empirical}, SimCLR (Simple Framework for Contrastive Learning)~\cite{chen2020simple}, and Barlow Twins~\cite{zbontar2021barlow}.}
In Section \ref{sec:exp:downstream}, we evaluate \our{}’s performance on downstream tasks such as classification, regression, object detection, and image retrieval\footnote{{We compare \our{} to several recently proposed methods and report their performance from the literature~\cite{xiao2020whatshouldnotbecontrastive,lee2021improving,chavhan2023amortised}, given that the code for \cite{xiao2020whatshouldnotbecontrastive} was not made available at the time of writing. As for the results of baseline SSL models and AugSelf \cite{lee2021improving}, we report their results from the literature except when our runs of those methods yielded results different by at least 2 pp. We mark such cases with $\dag$.}}. 
In Section \ref{sec:exp:cosine}, we analyze the sensitivity to augmentations of representations formed by \our{}. 
Next, we discuss the choice of hyperparameters of \our{} in Section \ref{sec:exp:ablation}. 
Furthermore, we refer to \ref{sec:exp:analysis} for a series of experiments that highlight several noteworthy features of \our{}, including but not limited to its effect on the minimization of contrastive loss and generalizing to augmentations that were not observed during pretraining.

In all experiments, unless specified otherwise, we utilize the ResNet-50 architecture~\cite{he2015deep} and conduct the self-supervised pretraining on ImageNet-100 - a 100-class subset of the ILSVRC dataset~\cite{russakovsky2015imagenet}  used commonly in the literature~\cite{tian2020makes,xiao2020whatshouldnotbecontrastive,lee2021improving,chavhan2023amortised}. We use the standard set of augmentations including horizontal flipping, random cropping, grayscaling, color jittering and Gaussian blurring~\cite{he2020momentum,lee2021improving,grill2020bootstrap}. For consistency in terms of hyperparameters, we follow~\cite{lee2021improving} for MoCo-v2
, and~\cite{chavhan2023amortised} for SimCLR. 
{We refer to \ref{sec:app:details} for the details of pretraining, evaluation, and implementation.}

\vspace{-0.25cm}
\subsection{Evaluation on downstream tasks}
\label{sec:exp:downstream}

\vspace{-0.25cm}

We begin the experimental analysis by addressing the most fundamental question -- how does \our{} impact the ability of models to generalize to downstream tasks? To answer it, we evaluate models pretrained via \our{} and other self-supervised techniques on a variety of downstream visual tasks, such as classification, regression, object detection, and image retrieval.

\vspace{-0.25cm}
\paragraph{Linear evaluation} We evaluate the performance of pretrained networks on the downstream tasks of classification and regression on 13 different datasets typically used for evaluation of SSL methods~\cite{chen2020simple,lee2021improving,chavhan2023amortised}, listed in Table \ref{tab:eval_datasets}.
We follow the linear evaluation protocol \cite{kornblith2019better,chen2020simple,lee2021improving}, described in detail in \ref{sec:app:eval}. We evaluate multiple self-supervised methods extended with \our{}, as well as other recently proposed extensions which increase sensitivity to augmentations~\cite{lee2021improving,xiao2020whatshouldnotbecontrastive,chavhan2023amortised} or prevent feature suppression in SSL~\cite{robinson2021contrastive}. 
We report the full results in Table \ref{tab:linear}. {We find that in the vast majority of cases, \our{} improves the performance of vanilla joint-embedding methods, as well as other SSL extensions that foster augmentation sensitivity~\cite{lee2021improving,chavhan2023amortised}.}

\begin{table*}[h]
  \caption{Linear evaluation on downstream classification and regression tasks. \our{} consistently improves representations formed by vanilla SSL approaches and performs better or comparably to other techniques of increasing sensitivity to augmentations \cite{xiao2020whatshouldnotbecontrastive,lee2021improving,chavhan2023amortised}.}
  \small
  \label{tab:linear}
  \centering
  \setlength\tabcolsep{2.5pt}
  \resizebox{\linewidth}{!}{
 \begin{tabular}{lccccccccccccc} 
    \toprule
    \textbf{Method} & C10 & C100 & Food & MIT & Pets & Flowers & Caltech & Cars & FGVCA & DTD & SUN & CUB & 300W  \\ 
    \toprule
   
    \multicolumn{14}{c}{\textit{SimCLR} \cite{chen2020simple}} \\
    \midrule

    Vanilla & 84.41$^\dag$	& 61.40	& 57.48$^\dag$	& 63.10$^\dag$ & 71.60$^\dag$ & 83.37$^\dag$ & \textbf{79.67}$^\dag$ & 35.14$^\dag$ & 40.03$^\dag$ & 64.90 & 46.92$^\dag$ & 30.98$^\dag$ & 88.59$^\dag$ \\ 
    AugSelf \cite{lee2021improving}$^\dag$ & 84.45	& 	62.67	& 	59.96	& 	63.21	& 	70.61	& 	\textbf{85.77}	& 	77.78	& 	37.38	& 	42.86	& 	65.53	& 	\textbf{49.18}	& 	34.24	& 	88.27 \\
    AI \cite{chavhan2023amortised} & 83.90 & 63.10 & -- & -- & 69.50 &	68.30 &	74.20 & -- & -- & 53.70 & -- & \textbf{38.60} & 88.00 \\
    \textbf{\our{}} & \textbf{86.31}	 & \textbf{64.36} & 	\textbf{60.67} & 	\textbf{63.96} & 	\textbf{72.33} & 	85.22 & 	79.62	 & \textbf{39.86} & 	\textbf{43.10} & 	\textbf{65.96} & 	48.91 & 	33.21	 & \textbf{88.88}\\
    \midrule
    
    \multicolumn{14}{c}{\textit{MoCo-v2} \cite{he2020momentum,chen2020improved}} \\
    \midrule
    Vanilla & 84.60 & 61.60 & 59.67 & 61.64 & 70.08 & 82.43 & 77.25 & 33.86 & 41.21 & 64.47 & 46.50 & 32.20 & $88.77^\dag$ \\
      AugSelf \cite{lee2021improving} & 85.26	& 63.90 &	60.78 &	63.36 &	73.46 &	85.70 &	78.93 &	37.35 &	39.47 &	66.22 &	48.52 & 37.00 &	$89.49^\dag$ \\
    AI \cite{chavhan2023amortised} & 81.30 &	64.60 &	-- & -- & \textbf{74.00} & 81.30 & 78.90 & -- & -- & \textbf{68.80} & -- & \textbf{41.40} & \textbf{90.00}	\\
      LooC \cite{xiao2020whatshouldnotbecontrastive} & -- & -- & -- & -- & -- & -- & -- & -- & -- & -- &  39.60 & -- & -- \\
      IFM~\cite{robinson2021contrastive}$^\dag$ & 83.36 & 60.22 & 59.86 & 60.60 & 72.99 & 85.73 & 78.77 & 36.54 & 41.05 & 62.34 & 47.48 & 35.90 & 88.92 \\
    \textbf{\our{}} & \textbf{86.32}	& \textbf{65.29}	& \textbf{61.93}		& \textbf{63.86}	& 	72.86	& 	\textbf{86.51}	& 	\textbf{79.63}	& 	\textbf{38.82}	& 	\textbf{42.03}		& 66.54		& \textbf{49.25}		& 36.22		& 88.93 \\
    \midrule

    \multicolumn{14}{c}{\textit{Barlow Twins} \citep{zbontar2021barlow}} \\
    \midrule

    Vanilla$^\dag$ & 85.90	& 66.10	& 59.41	& 61.72	& 72.30	& 87.13	& {81.95}	& {41.54} & 44.40	& 65.85	& 49.18 & 35.02	& 89.04  \\
    AugSelf \citep{lee2021improving}$^\dag$ & \textbf{87.28} & 66.98	& 60.52	& 63.96	& 72.11	& 86.68	& 81.73 & 39.88 & 44.23	& 65.21	& 47.71	& 37.02 & 88.88 \\
    \textbf{\our{}} & 87.03 &	\textbf{67.27} &	\textbf{62.19} & \textbf{65.08}	& \textbf{72.75}	& \textbf{87.99}	& \textbf{82.56}	& \textbf{41.68}	& \textbf{46.63} & 	\textbf{66.31} & 	\textbf{50.09} & 	\textbf{38.25}	& \textbf{89.52} \\
    \midrule
    \multicolumn{14}{c}{{\textit{MoCo-v3} \cite{he2020momentum,chen2021empirical} with ViT-Small \cite{dosovitskiy2021image} pretrained on ImageNet-1K.}} \\
    \midrule
    Vanilla$^\dag$ & 83.17	& 62.40	& 56.15	& 53.28 & 62.29 & 81.48	& 69.63	& 28.63	& 32.84	& 57.18	& 42.16	& 35.00 & 87.42 \\
    AugSelf \cite{lee2021improving}$^\dag$ & {84.25} & 64.12 & \textbf{58.28} & \textbf{56.12} & \textbf{63.93} & \textbf{83.13} & 72.45 & 29.64 & 32.54& \textbf{60.27} & {43.22} & \textbf{37.16} & 87.85 \\ 
    \textbf{\our{}} & \textbf{85.13} & \textbf{64.67} & 57.30 & 55.90 & 63.88 & 82.42 & \textbf{73.53} & \textbf{30.92} & \textbf{35.91} & 58.24 & \textbf{43.37} & 36.09 & \textbf{88.53}  \\

    \bottomrule
  \end{tabular}
  } 
\end{table*}

\vspace{-0.25cm}

\paragraph{Object detection} 
{We next evaluate the pretrained networks on a more challenging task of object detection on the VOC (Visual Object Classification) 2007 dataset \cite{everingham2007pascalvoc}.} We follow the training scheme of \cite{he2020momentum,chen2020improved}, except that we only train the object detector modules and keep the feature extractor parameters fixed during training for detection to better compare the pretrained representations. We report the Average Precision (AP) \cite{tsung2014cocodataset} of models pretrained through MoCo-v2 and SimCLR \cite{chen2020simple} with AugSelf \cite{lee2021improving} and \our{} extensions in Table \ref{tab:detection}. The compared approaches yield similar results, with \our{} representation surpassing the vanilla methods and AugSelf. 
\begin{wraptable}{r}{0.5\textwidth}
    \vspace{0.75cm}
  \caption{Average Precision of object detection on VOC dataset \cite{everingham2007pascalvoc,tsung2014cocodataset}. \our{} extension of MoCo-v2 and SimCLR outperforms the vanilla approaches and AugSelf extension by a small margin.}
  \label{tab:detection}
  \centering
    \resizebox{\linewidth}{!}{
  \begin{tabular}{lcc} %
    \toprule
   \textbf{Method} & \textit{MoCo-v2~\cite{he2020momentum,chen2020improved}}  & \textit{SimCLR~\cite{chen2020simple}} \\      
    \midrule
    Vanilla & 45.12 & 44.74 \\
    AugSelf \cite{lee2021improving} & 45.20 & 44.50 \\
    \textbf{\our{}} & \textbf{45.90} & \textbf{45.60} \\    
    \bottomrule
  \end{tabular}
  }
  \vspace{-0.5cm}
\end{wraptable}

\begin{figure}[h]
    \centering
    \begin{subfigure}[t]{0.48\textwidth}
        \centering
            \includegraphics[width=\textwidth]{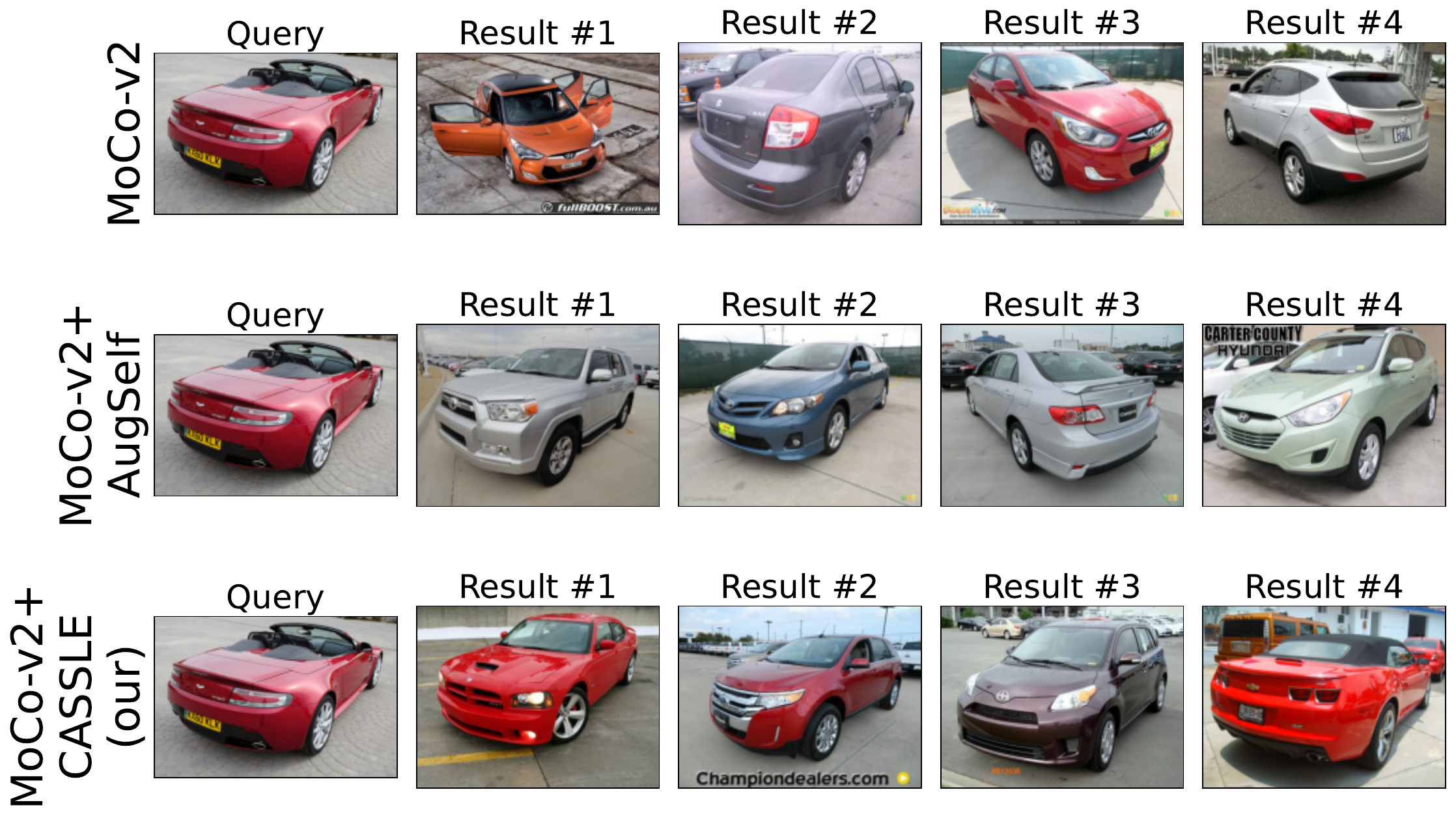} \\
        \includegraphics[width=\textwidth]{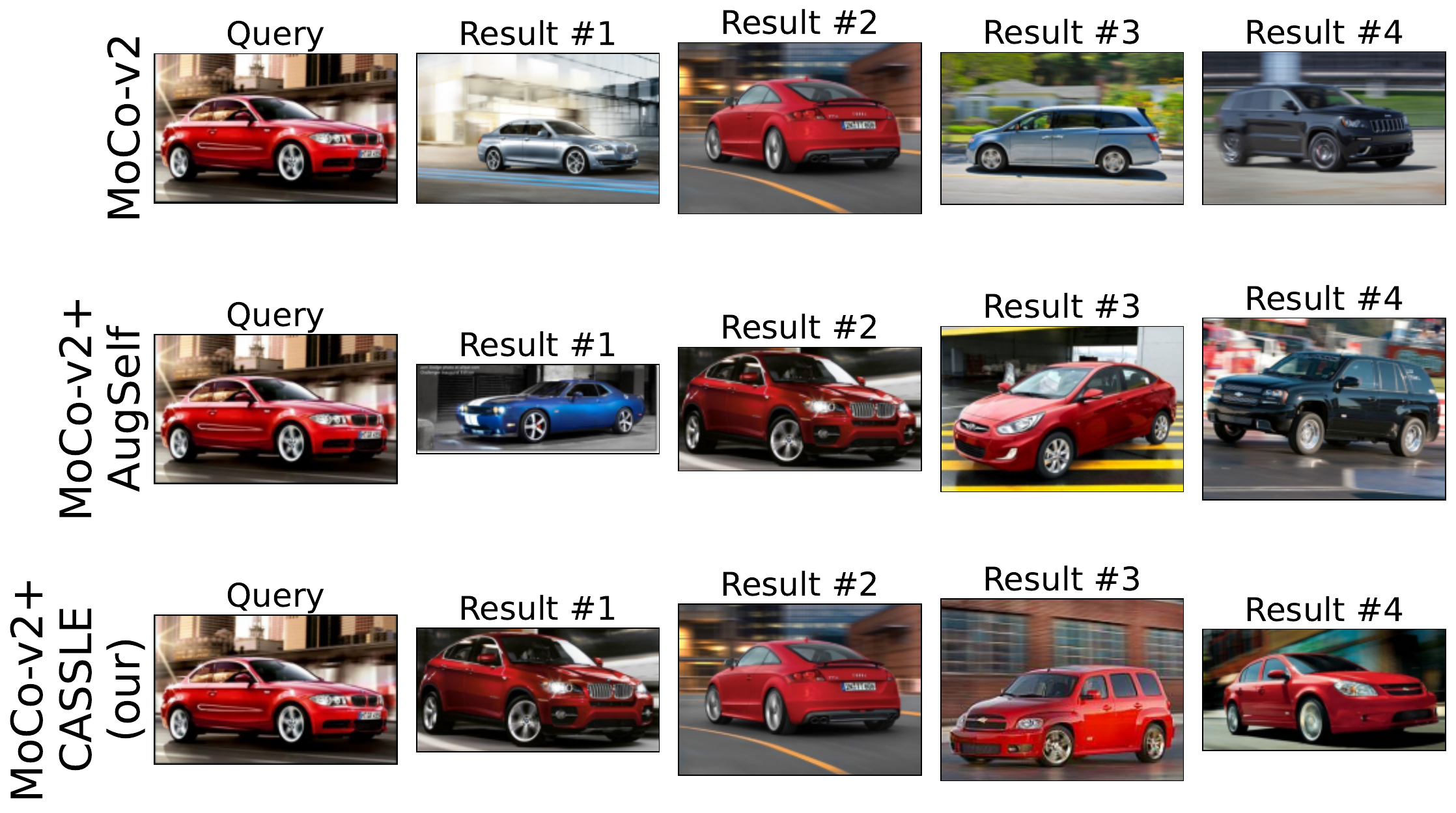} \\

        \includegraphics[width=\textwidth]{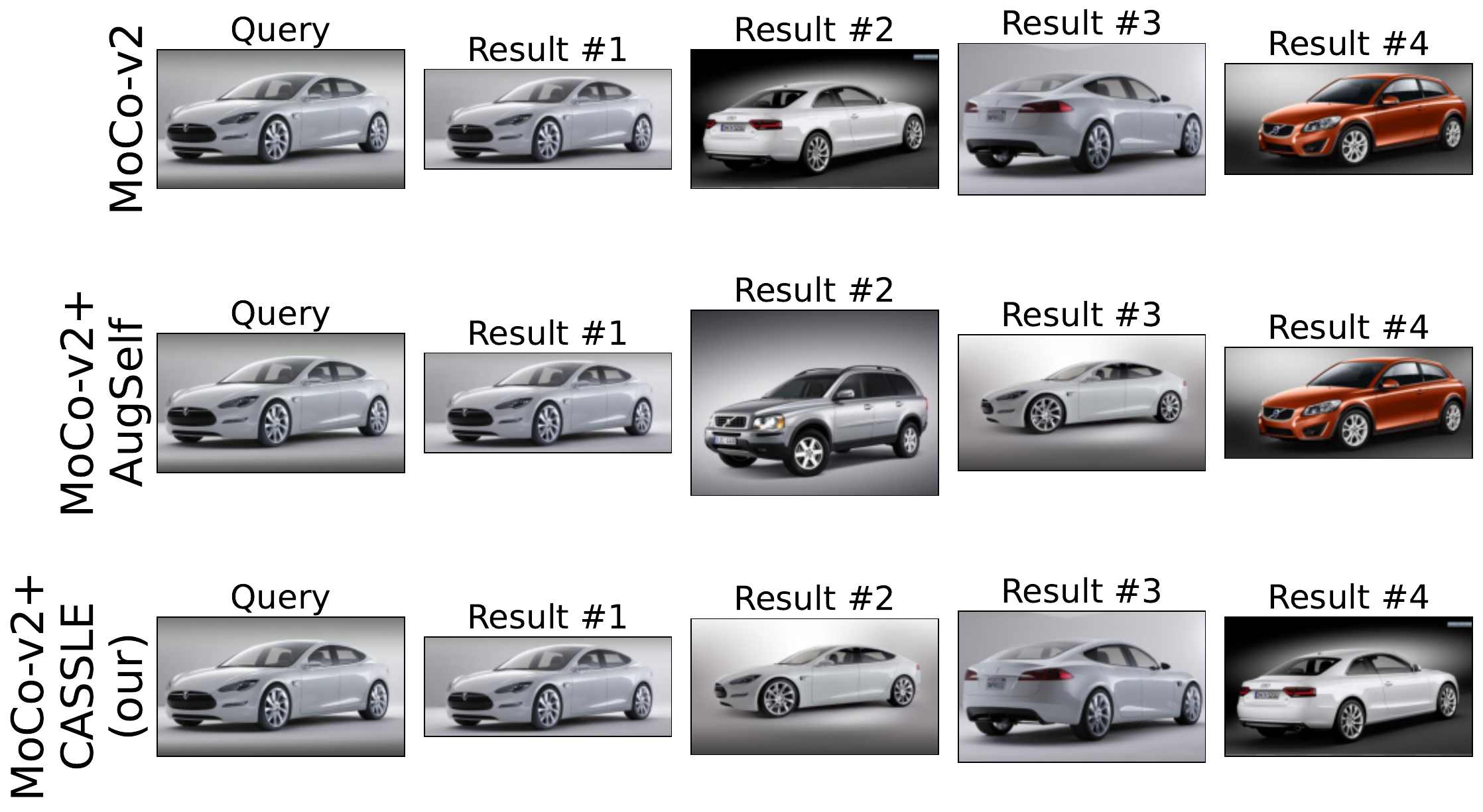}
            \caption{Cars}
    \end{subfigure}
    \hfill
    \begin{subfigure}[t]{0.48\textwidth}
        \centering
        \includegraphics[width=\textwidth]{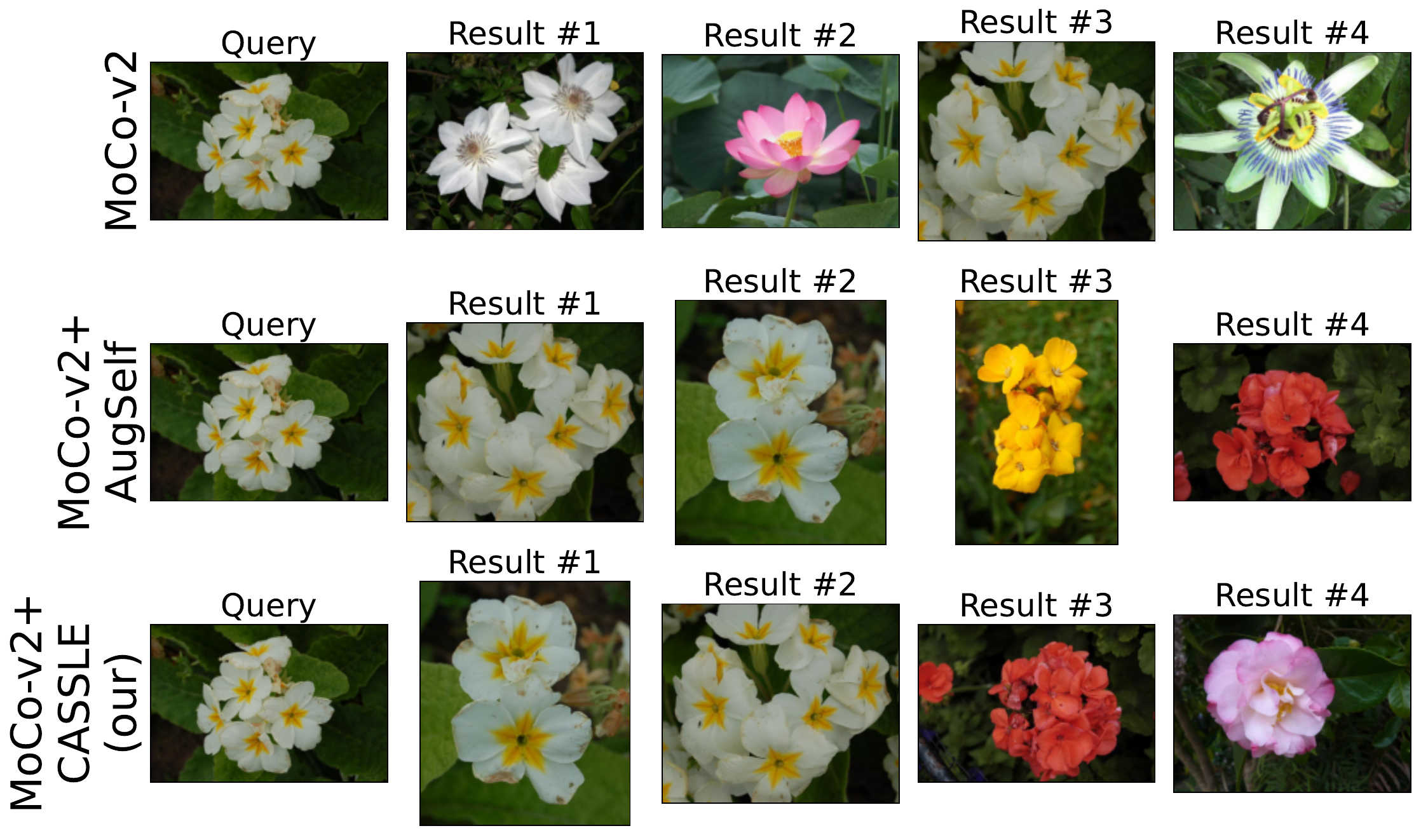} \\

        \includegraphics[width=\textwidth]{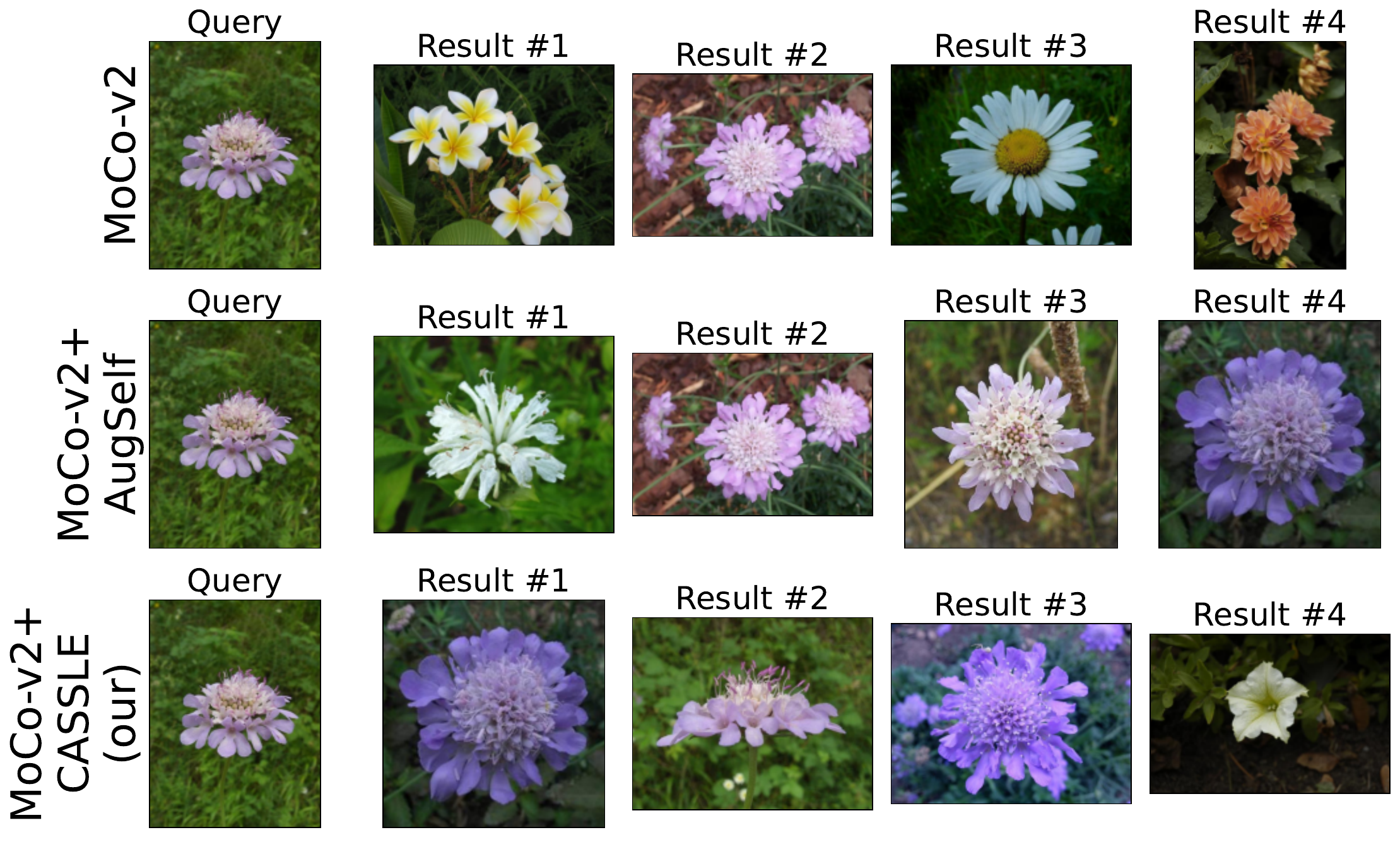} \\
        \includegraphics[width=\textwidth]{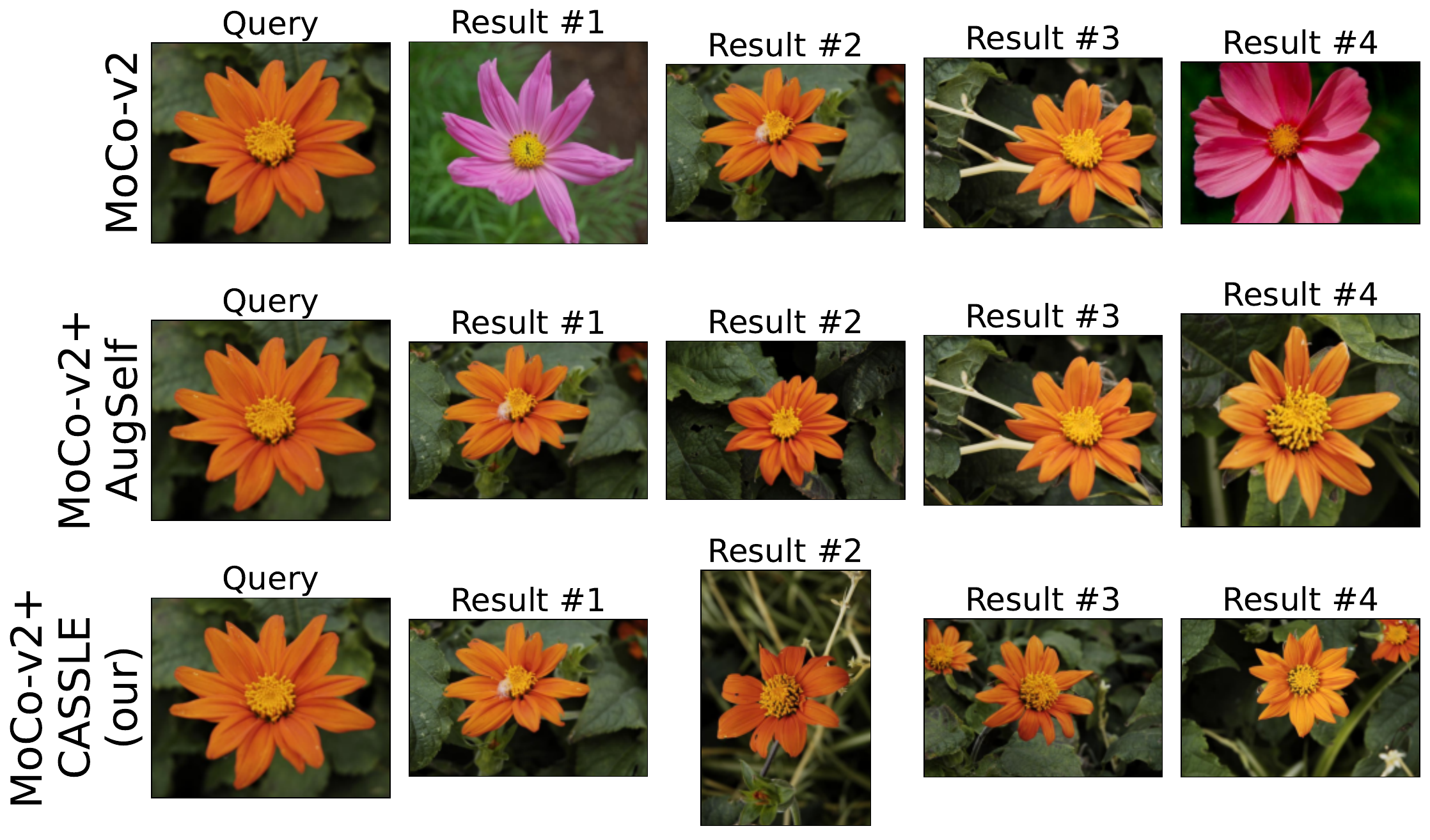} \\
        \caption{Flowers}
    \end{subfigure}
    \vspace{-0.15cm}
    \caption{Image retrieval examples for Cars and Flowers datasets.}
    \label{app:fig:exp_nn}
    \vspace{-0.7cm}
\end{figure}

\paragraph{Image retrieval} 
Finally, we evaluate the pretrained models on the task of image retrieval. We gather the features of images from the Cars and Flowers test sets and for a given query image, select four images closest to it in terms of the cosine distance of final feature extractor representations. 
We compare the images retrieved by MoCo-v2, AugSelf \cite{lee2021improving} and \our{} in Figure \ref{app:fig:exp_nn}. \our{} selects pictures of cars that are the most consistent in terms of color. In the case of flowers, the nearest neighbor retrieved by the vanilla model is a different species than that of the first query image, whereas both \our{} and AugSelf select the first two nearest neighbors from the same species but then retrieve images of flowers with similar shapes, but different colors. This again indicates greater reliability of features learned by \our{}. For subsequent queries, \our{} and AugSelf retrieve in general more consistently looking images, in particular in terms of color scheme. This indicates a greater sensitivity of those models to shifts in colors.

\newpage
\subsection{Analysis of representations formed by \our{}}
\label{sec:exp:cosine}
\begin{figure}[h]
    \centering
\includegraphics[width=\textwidth]{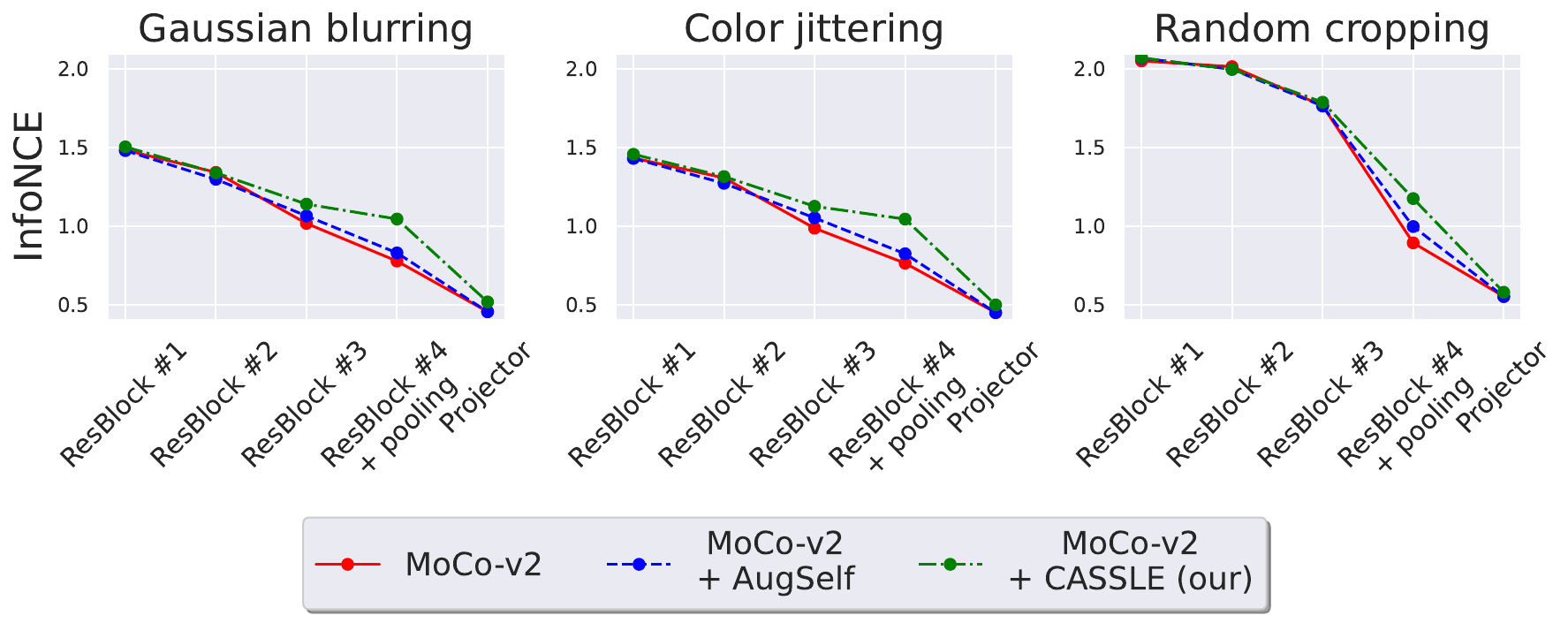}
    \caption{A comparison of InfoNCE loss measured on different kinds of augmentations at subsequent stages of the ResNet-50 and projectors pretrained by vanilla, AugSelf \cite{lee2021improving} and \our{} variants of MoCo-v2. Feature extractor representation of \our{} yields higher InfoNCE values which suggests that it is more susceptible to augmentations.
    }
    \label{fig:exp_invariance}
    \vspace{-0.5cm}
\end{figure}

\vspace{-0.25cm}
{We have demonstrated that \our{} has a positive impact on the quality of trained representations. We now investigate the differences in those representations. Specifically, we investigate the awareness of augmentation-induced data perturbations in the intermediate and final representations of pretrained networks.}
As a proxy metric for measuring this, we choose the InfoNCE loss \cite{oord2019representation,chen2020simple}.
The value of InfoNCE is high if embeddings of pairs of augmented images are less similar to one another than to embeddings of other images, and low if positive embedding pairs are matched correctly, and thus, the given representation is invariant to augmentations. We report the mean InfoNCE loss values for different augmentation types at subsequent stages of ResNet-50 and projectors of vanilla MoCo-v2, AugSelf~\cite{lee2021improving} and \our{} in Figure \ref{fig:exp_invariance}.

In all networks, the augmentation awareness decreases gradually throughout the feature extractor and projector stages. In \our{}, we observe a much softer decline in the feature extractor stages and a sharper one in the projector. Representations of \our{} feature extractor are on average more difficult to match together than those of vanilla MoCo-v2 and AugSelf~\cite{lee2021improving}. This implies that the \our{} feature extractor is indeed more sensitive to augmentations than its counterparts. On the other hand, representations of all projectors, including \our{}, are similarly separable. {This suggests that the conditioning mechanism helps \our{} projector to better amortize the augmentation-induced differences between the embeddings produced by the feature extractor.}

The above observations indicate that in the vanilla and (to a slightly lesser extent) AugSelf approaches, both the projector and the intermediate representations are enforced to be augmentation-invariant. 
On the other hand, in \our{}, the task of augmentation invariance is solved to a larger degree by the projector, and to a smaller degree by the feature extractor, allowing it to be more augmentation-aware. 
As shown in Section \ref{sec:exp:downstream}, this sensitivity does not prevent the \our{} feature extractor from achieving similar or better performance than its counterparts when transferred to downstream tasks.

\subsection{Ablation study}
\label{sec:exp:ablation}

We have demonstrated the quality and properties of representations learned We next examine the impact of different hyperparameters of \our{}.
We compare different variants of 
MoCo-v2+\our{} on the same classification and regression tasks as in Section \ref{sec:exp:downstream}. We rank the models from best to worst performance on each task and report the average ranks in Table \ref{tab:ablation}. 

\begin{table*}[h]
  \caption{Ablation study of \our{} parameters. The results are computed with MoCo-v2+\our{}. It is best to condition \our{} on all available augmentation information. \our{} yields the best results when implemented by concatenating or adding the augmentation and image embeddings together.}
  \label{tab:ablation}
  \centering
  \setlength\tabcolsep{2.5pt}
    \resizebox{\linewidth}{!}{

 \begin{tabular}{ccccccccccccccc} 
    \toprule
     \textbf{Parameter} & C10 & C100 & Food & MIT & Pets & Flowers & Caltech & Cars & FGVCA & DTD & SUN & CUB & 300W & \textbf{Avg. rank $\downarrow$}\\
    \toprule
    \multicolumn{13}{l}{Augmentation information contents} \\
    \midrule
    $\omega^{\{c\}}$  & 84.89 & 62.95 & 59.74 & 63.96 & 72.26 & 83.55 & 79.66 & 38.78 & 42.03 & 65.11 & 48.44 & 33.86 & 89.17 & 4.54 \\
    $\omega^{\{c,j\}}$ & 85.56 & 64.26 & 60.35 & 62.61 & 71.97 & 84.73 & 79.86 & 38.13 & 42.17 & 66.28 & 48.01 & 34.24& 88.76 & 4.54 \\
    $\omega^{\{c,j,d\}}$ & 85.87 & 63.91 & 61.07 & 63.51 & 72.71 & 86.53 & 79.51 & 38.27 & 42.53 & 66.70 & 49.27 & 35.76 & 89.11 & 3.00 \\
    $\omega^{\{c,j,b,f\}}$ & 86.16 & 64.51 & 60.80 & 63.81 & 72.83 & 84.66 & 79.90 & 38.93 & 43.02 & 66.12 & 48.96 & 34.40 & 88.69  & 2.84 \\
    $\omega^{\{c,j,b,f,g\}}$ & 85.85 & 64.14 & 61.24 & 63.73 & 72.88 & 84.50 & 79.93 & 38.23 & 41.28 & 65.27 & 48.90 & 34.47 & 88.78 & 3.53 \\
     $\omega^{\{c,j,b,f,g,d\}}$ & 86.99 & 65.28 & 61.83 & 63.51 & 73.22 & 86.55 & 79.87 & 37.97 & 41.70 & 67.18 & 48.85 & 36.92 & 89.03 & \textbf{2.46} \\
     \midrule
    \multicolumn{13}{l}{{Impact of utilizing color difference information during pretraining}} \\
    \midrule
    {\bf \our{} ${\omega}^{\{c,j\}}$} & 85.56 & 64.26 &  60.35 & 62.61 & 71.97 & 84.73 & 79.86 & 38.13 & 42.17 & 66.28 &  48.01 &  34.24 & 88.76 & 4.53 \\
    \textbf{AugSelf $\omega^{\{c,j\}}$} & 85.26	& 63.90 &	60.78 &	63.36 &	 {73.46} &	 {85.70} &	78.93 &	37.35 &	39.47 &	 {66.22} &	48.52 &  {37.00} &	{89.49} & 4.23 \\
    {\bf \our{} $\omega^{\{c,j,d\}}$} & 85.87 & 63.91 & 61.07 & 63.51 & 72.71 & 86.53 & 79.51 & 38.27 & 42.53 & 66.70 & 49.27 & 35.76 & 89.11 & 2.76 \\
    {\bf AugSelf $\omega^{\{c,j,d\}}$} & 84.95 & 64.06 & 61.53 & 63.06& { {73.52}} & 86.25 & 77.38 & 36.00& {42.54} & 66.33 & 48.65 & {37.40} & 88.36 & 3.69 \\
    {\bf \our{}} $\omega^{\{c,j,b,f,g\}}$ &  {85.85} &  {64.14} & {61.24} & {63.73} & 72.88 & 84.50 & {{79.93}} &  {38.23} &  {41.28} & 65.27 & {48.90} & 34.47 & 88.78 & 3.46 \\
    {\bf \our{} $\omega^{\{c,j,b,f,g,d\}}$} & 86.99 & 65.28 & 61.83 & 63.51 & 73.22 & 86.55 & 79.87 & 37.97 & 41.70 & 67.18 & 48.85 & 36.92 & 89.03 & \textbf{2.23} \\
    \midrule
    \multicolumn{13}{l}{Method of conditioning the projector} \\
    \midrule
    \textbf{Concatenation}  & 86.99 & 65.28 & 61.83 & 63.51 & 73.22 & 86.55 & 79.87 & 37.97 & 41.70 & 67.18 & 48.85 & 36.92 & 89.03 & \textbf{1.92} \\
    \textbf{Addition}  & 86.45 & 65.40 & 63.00 & 65.15 & 71.34 & 86.91 & 79.79 & 37.83 & 42.18 & 66.17 & 49.28 & 37.42 & 88.87 & \textbf{1.92} \\
    \textbf{Multiplication}  & 86.72 & 66.70 & 60.65 & 60.97 & 64.60 & 85.17 & 80.09 & 33.54 & 41.56 & 63.99 & 47.63 & 32.15 & 89.48 & 2.69 \\
    \textbf{Hypernetwork} & 84.70 & 63.55 & 60.62 & 64.10 & 67.16 & 82.76 & 78.47 & 33.39 & 39.85 & 66.44 & 47.43 & 30.48 & 89.11 &  3.46 \\
    \midrule
    \multicolumn{13}{l}{Depth of the \guidingnet{} (0 denotes concatenating
raw $\mathbf{\omega}$ to $\mathbf{e}$)} \\
    \midrule
    \textbf{0} & 86.53 & 65.99 & 62.54 & 61.72 & 69.04 & 85.46 & 80.74 & 36.44 & 41.91 & 65.64 & 48.55 & 33.88 & 92.72 & 3.38\\
    \textbf{2} & 86.57 & 64.80 & 61.85 & 62.68 & 72.79 & 86.31 & 79.93 & 37.85 & 42.87 & 66.32 & 49.23 & 35.39 & 88.86 & 3.08 \\
    \textbf{4}  & 86.32 & 65.16 & 61.98 & 64.93 & 72.64 & 86.49 & 79.75 & 38.64 & 41.46 & 66.91 & 49.71 & 36.56 & 89.27 & 2.38 \\
    \textbf{6}  & 86.99 & 65.28 & 61.83 & 63.51 & 73.22 & 86.55 & 79.87 & 37.97 & 41.70 & 67.18 & 48.85 & 36.92 & 89.03 & \textbf{2.15} \\
    \textbf{8}  & 85.47 & 64.88 & 61.49 & 63.13 & 72.41 & 85.65 & 78.22 & 37.82 & 40.71 & 66.70 & 49.21 & 36.09 & 88.90 & 4.00 \\
    \midrule 
    \multicolumn{13}{l}{Hidden size of the \guidingnet{}} \\
    \midrule
    \textbf{16} &  84.51	& 63.40	& 61.38	& 62.31	& 71.61	& 85.40	& 78.96	& 37.53	& 41.84	& 66.54	& 48.68	& 35.54	& 88.78 & 5.54  \\
    \textbf{32} & 85.43	& 63.94	& 61.93 & 64.18	& 72.05	& 85.67 & 79.80 & 37.86 & 41.52 & 67.13 & 48.71 & 35.64 & 88.67 & 4.38 \\
    \textbf{64} & 86.99 & 65.28 & 61.83 & 63.51 & 73.22 & 86.55 & 79.87 & 37.97 & 41.70 & 67.18 & 48.85 & 36.92 & 89.03 & \textbf{2.54}  \\
    \textbf{128}  & 86.24 & 64.66 & 61.95 & 64.25 & 72.12 & 86.72 & 79.56 & 37.79 & 42.71 & 67.61 & 49.52 & 36.87 & 88.99 & 2.62 \\
    \textbf{256} & 86.23 & 	65.63 & 61.77 & 61.79 & 72.03 & 85.69 & 80.38 & 37.86 & 40.94 & 67.07 & 49.59 & 37.00 & 89.37 & 3.31 \\
    \textbf{512}  & 85.77 & 66.05 & 62.21 & 64.55 & 72.45 & 86.38 & 80.06 & 37.91 & 41.94 & 66.22 & 49.35 & 36.24 & 88.87 & 2.69 \\    
    
    \bottomrule
    
  \end{tabular}
  }
  \vspace{-0.5cm}
\end{table*}

\paragraph{Augmentation information contents} We compare conditioning the projector with different subsets of augmentation information\footnote{
We recall the notation used for Augmentation information contents -- $\omega^{\{x,y\}}$ denotes including parameters of augmentations $\{x,y\}$ in augmentation information vector $\omega$. For example, $\omega^{\{c,j\}}$ denotes $\omega$ containing \textbf{c}ropping and color \textbf{j}ittering parameters.}
The average best representation is trained with conditioning on all possible augmentation information. Moreover, using the additional \textbf{color difference} ($\omega^d$) information additionally improves the results, indicating that it is indeed useful to consider not only augmentation parameters but also information about its effects.

\paragraph{Impact of utilizing color difference information} We verify that the improved performance of \our{} does not stem solely from using augmentation information that has not been considered in prior works, i.e. color difference $(\omega^d)$. We compare a variant of AugSelf which learns to predict color difference values in addition to augmentation information used by~\cite{lee2021improving}, i.e. cropping ($\omega^c$) and color jittering ($\omega^j$), as well as a variant of \our{} conditioned on all augmentation information \emph{except} $\omega^d$ (i.e. $\omega^{\{c,j,b,f,g\}}$). We find that while including $\omega^d$ improves the performance of AugSelf and \our{}, both variants of \our{} achieve better results than both variants of AugSelf.
{We also compare variants \our{} conditioned with the same parameters as used with AugSelf, i.e. $\omega^{\{c,j\}}$ and $\omega^{\{c,j,d\}}$. Models that use only the jittering and cropping information ($\omega^{\{c,j\}}$) perform similarly, achieving the mean ranks of 4.53 and 4.23. In the case of $\omega^{\{c,j,d\}}$, \our{} and AugSelf differ by a larger margin, achieving the mean rank of 2.76, and 3.69 respectively. The above results underscore the usefulness of using all available augmentation parameters, including color difference, in \our{}. In contrast, Lee et. al.~\cite{lee2021improving} observe that predicting all augmentation parameters in AugSelf lowers the quality of the trained model.}

\paragraph{Method of conditioning the projector} 
We compare conditioning the projector through (i) concatenation, element-wise (ii) addition or (iii) multiplication, or (iv) using augmentation information as an input to a hypernetwork~\cite{ha2016hypernetworks} which generates the parameters of $\pi$. Conditioning through \textbf{concatenation} and \textbf{addition} yields on average the strongest performance on downstream tasks. We choose to utilize the \textbf{concatenation} method in our experiments, as it requires a slightly smaller \guidingnet{}.

\paragraph{Size of the \guidingnet{}} While \our{} is robust to the size of the $\gamma$ MLP, using the depth and hidden size of 6 and 64, respectively, yields the strongest downstream performance. In particular, the variant of \our{} that utilizes the \guidingnet{} performs better than the variant that concatenates $\mathbf{e}$ to raw augmentation embeddings $\mathbf{\omega}$. Given such an architecture of the \guidingnet{}, our computation overhead is negligible as we increase the overall number of parameters by around $0.1\%$ compared to vanilla SSL approaches.

\section{Conclusion}
\label{sec:conclusion}
\vspace{-0.25cm}
{In this paper, we propose \our{}: a novel method for augmentation-aware self-supervised learning that retains information about data augmentations in the representation space. To accomplish this, we introduce the concept of the 
conditioned
projector, which receives augmentation information while processing the representation vector.} Our solution necessitates only small architectural changes and no additional auxiliary loss components. Therefore, the training concentrates on contrastive loss, which enhances overall performance.
We compare our solution with existing augmentation-aware SSL methods and demonstrate its superior performance on downstream tasks, particularly when augmentation invariance leads to the loss of vital information. 
Moreover, we show that it obtains representations more sensitive to augmentations than the baseline methods.
Overall, our method offers a straightforward and efficient approach to retaining information about data augmentations in the representation space. It can be directly applied to SSL methods, contributing to the further advancement of augmentation-aware self-supervised learning.

\newpage
\section*{Acknowledgments}
This research has been supported by the flagship project entitled "Artificial Intelligence Computing Center Core Facility" from the Priority Research Area DigiWorld under the Strategic Programme Excellence Initiative at Jagiellonian University, and by the Horizon Europe Programme (HORIZON-CL4-2022-HUMAN-02) under the project "ELIAS: European Lighthouse of AI for Sustainability", GA no. 101120237.
The research of Marcin Przewięźlikowski was supported by the National Science Centre (Poland), grant no. 2023/49/N/ST6/03268. The research of Marek Śmieja was supported by the National Science Centre (Poland), grant no. 2022/45/B/ST6/01117. The research of Bartosz Zieliński is funded in part by National Science Centre (Poland) grant number 2022/47/B/ST6/03397. Bartłomiej Twardowski acknowledges the grant RYC2021-032765-I. We gratefully acknowledge Polish high-performance computing infrastructure PLGrid (HPC Centers: ACK Cyfronet AGH) for providing computer facilities and support within computational grant no. PLG/2023/016303.

The authors would like to express their gratitude to Przemysław Spurek, Tomasz Trzciński, Maciej Wołczyk, Michał Zając, Marcin Sendera, Filip Szatkowski, and Daniel Marczak for the insightful discussions throughout the work. The authors would also like to thank Hankook Lee and Ruchika Chavhan for releasing the codebases of AugSelf and Amortised Invariance, which served as bases for the development of \our{}.


\bibliographystyle{plain}
{\small
\bibliography{ref.bib}

\begin{thebibliography}{10}

\bibitem{saleh2022selfsupervisedsurvey}
Saleh Albelwi.
\newblock Survey on self-supervised learning: Auxiliary pretext tasks and contrastive learning methods in imaging.
\newblock {\em Entropy}, 24(4), 2022.

\bibitem{bai2021selfsupervised}
Haoping Bai, Meng Cao, Ping Huang, and Jiulong Shan.
\newblock Self-supervised semi-supervised learning for data labeling and quality evaluation.
\newblock In {\em NeurIPS Workshop}, 2021.

\bibitem{balestriero2023cookbook}
Randall Balestriero, Mark Ibrahim, Vlad Sobal, Ari~S. Morcos, Shashank Shekhar, Tom Goldstein, Florian Bordes, Adrien Bardes, Gr{\'e}goire Mialon, Yuandong Tian, Avi Schwarzschild, Andrew~Gordon Wilson, Jonas Geiping, Quentin Garrido, Pierre Fernandez, Amir Bar, Hamed Pirsiavash, Yann LeCun, and Micah Goldblum.
\newblock A cookbook of self-supervised learning.
\newblock {\em ArXiv}, abs/2304.12210, 2023.

\bibitem{bardes2022vicreg}
Adrien Bardes, Jean Ponce, and Yann LeCun.
\newblock {VICR}eg: Variance-invariance-covariance regularization for self-supervised learning.
\newblock In {\em International Conference on Learning Representations}, 2022.

\bibitem{becker1992self}
Suzanna Becker and Geoffrey Hinton.
\newblock Self-organizing neural network that discovers surfaces in random-dot stereograms.
\newblock {\em Nature}, 355:161--3, 02 1992.

\bibitem{Bengar2021ICCV}
Javad~Zolfaghari Bengar, Joost van~de Weijer, Bartlomiej Twardowski, and Bogdan Raducanu.
\newblock Reducing label effort: Self-supervised meets active learning.
\newblock In {\em Proceedings of the IEEE/CVF International Conference on Computer Vision (ICCV) Workshops}, pages 1631--1639, October 2021.

\bibitem{bhardwaj2023steerable}
Sangnie Bhardwaj, Willie McClinton, Tongzhou Wang, Guillaume Lajoie, Chen Sun, Phillip Isola, and Dilip Krishnan.
\newblock Steerable equivariant representation learning, 2023.

\bibitem{bordes2022guillotine}
Florian Bordes, Randall Balestriero, Quentin Garrido, Adrien Bardes, and Pascal Vincent.
\newblock Guillotine regularization: Why removing layers is needed to improve generalization in self-supervised learning.
\newblock {\em Transactions on Machine Learning Research}, 2023.

\bibitem{bossard2014food101}
Lukas Bossard, Matthieu Guillaumin, and Luc Van~Gool.
\newblock Food-101 -- mining discriminative components with random forests.
\newblock In {\em European Conference on Computer Vision}, 2014.

\bibitem{gpt}
Tom Brown, Benjamin Mann, Nick Ryder, Melanie Subbiah, Jared~D Kaplan, Prafulla Dhariwal, Arvind Neelakantan, Pranav Shyam, Girish Sastry, Amanda Askell, et~al.
\newblock Language models are few-shot learners.
\newblock volume~33, pages 1877--1901, 2020.

\bibitem{caron2020unsupervised}
Mathilde Caron, Ishan Misra, Julien Mairal, Priya Goyal, Piotr Bojanowski, and Armand Joulin.
\newblock Unsupervised learning of visual features by contrasting cluster assignments.
\newblock In H.~Larochelle, M.~Ranzato, R.~Hadsell, M.F. Balcan, and H.~Lin, editors, {\em Advances in Neural Information Processing Systems}, volume~33, pages 9912--9924. Curran Associates, Inc., 2020.

\bibitem{caron2021emerging}
Mathilde Caron, Hugo Touvron, Ishan Misra, Herv\'e J\'egou, Julien Mairal, Piotr Bojanowski, and Armand Joulin.
\newblock Emerging properties in self-supervised vision transformers.
\newblock In {\em Proceedings of the International Conference on Computer Vision (ICCV)}, 2021.

\bibitem{chavhan2023amortised}
Ruchika Chavhan, Jan Stuehmer, Calum Heggan, Mehrdad Yaghoobi, and Timothy Hospedales.
\newblock Amortised invariance learning for contrastive self-supervision.
\newblock In {\em The Eleventh International Conference on Learning Representations}, 2023.

\bibitem{chen2020simple}
Ting Chen, Simon Kornblith, Mohammad Norouzi, and Geoffrey Hinton.
\newblock A simple framework for contrastive learning of visual representations.
\newblock In Hal~Daumé III and Aarti Singh, editors, {\em Proceedings of the 37th International Conference on Machine Learning}, volume 119 of {\em Proceedings of Machine Learning Research}, pages 1597--1607. PMLR, 13--18 Jul 2020.

\bibitem{chen2021intriguing}
Ting Chen, Calvin Luo, and Lala Li.
\newblock Intriguing properties of contrastive losses.
\newblock In A.~Beygelzimer, Y.~Dauphin, P.~Liang, and J.~Wortman Vaughan, editors, {\em Advances in Neural Information Processing Systems}, 2021.

\bibitem{chen2020improved}
Xinlei Chen, Haoqi Fan, Ross~B. Girshick, and Kaiming He.
\newblock Improved baselines with momentum contrastive learning.
\newblock {\em CoRR}, abs/2003.04297, 2020.

\bibitem{chen2021exploring}
Xinlei Chen and Kaiming He.
\newblock Exploring simple siamese representation learning.
\newblock In {\em Proceedings of the IEEE/CVF Conference on Computer Vision and Pattern Recognition (CVPR)}, pages 15750--15758, June 2021.

\bibitem{chen2021empirical}
Xinlei Chen, Saining Xie, and Kaiming He.
\newblock An empirical study of training self-supervised vision transformers.
\newblock In {\em Proceedings of the IEEE/CVF International Conference on Computer Vision (ICCV)}, pages 9640--9649, October 2021.

\bibitem{cimpoi14describing}
M.~Cimpoi, S.~Maji, I.~Kokkinos, S.~Mohamed, , and A.~Vedaldi.
\newblock Describing textures in the wild.
\newblock In {\em Proceedings of the {IEEE} Conf. on Computer Vision and Pattern Recognition ({CVPR})}, 2014.

\bibitem{devlin2019bert}
Jacob Devlin, Ming-Wei Chang, Kenton Lee, and Kristina Toutanova.
\newblock {BERT}: Pre-training of deep bidirectional transformers for language understanding, June 2019.

\bibitem{doersch2015unsupervised}
Carl Doersch, Abhinav Gupta, and Alexei~A. Efros.
\newblock Unsupervised visual representation learning by context prediction.
\newblock In {\em Proceedings of the IEEE International Conference on Computer Vision (ICCV)}, December 2015.

\bibitem{dosovitskiy2021image}
Alexey Dosovitskiy, Lucas Beyer, Alexander Kolesnikov, Dirk Weissenborn, Xiaohua Zhai, Thomas Unterthiner, Mostafa Dehghani, Matthias Minderer, Georg Heigold, Sylvain Gelly, Jakob Uszkoreit, and Neil Houlsby.
\newblock An image is worth 16x16 words: Transformers for image recognition at scale.
\newblock In {\em International Conference on Learning Representations}, 2021.

\bibitem{ericsson2022selfsupervised}
Linus Ericsson, Henry Gouk, and Timothy Hospedales.
\newblock Why do self-supervised models transfer? on the impact of invariance on downstream tasks.
\newblock In {\em 33rd British Machine Vision Conference 2022, {BMVC} 2022, London, UK, November 21-24, 2022}. {BMVA} Press, 2022.

\bibitem{everingham2007pascalvoc}
M.~Everingham, L.~Van~Gool, C.~K.~I. Williams, J.~Winn, and A.~Zisserman.
\newblock The {PASCAL} {V}isual {O}bject {C}lasses {C}hallenge 2007 {(VOC2007)} {R}esults.
\newblock http://www.pascal-network.org/challenges/VOC/voc2007/workshop/index.html.

\bibitem{li2006caltech}
Li~Fei-Fei, R.~Fergus, and P.~Perona.
\newblock One-shot learning of object categories.
\newblock {\em IEEE Transactions on Pattern Analysis and Machine Intelligence}, 28(4):594--611, 2006.

\bibitem{garrido2022duality}
Quentin Garrido, Yubei Chen, Adrien Bardes, Laurent Najman, and Yann Lecun.
\newblock On the duality between contrastive and non-contrastive self-supervised learning.
\newblock 2022.

\bibitem{garrido2023self}
Quentin Garrido, Laurent Najman, and Yann Lecun.
\newblock Self-supervised learning of split invariant equivariant representations.
\newblock In Andreas Krause, Emma Brunskill, Kyunghyun Cho, Barbara Engelhardt, Sivan Sabato, and Jonathan Scarlett, editors, {\em Proceedings of the 40th International Conference on Machine Learning}, volume 202 of {\em Proceedings of Machine Learning Research}, pages 10975--10996. PMLR, 23--29 Jul 2023.

\bibitem{gidaris2018unsupervised}
Spyros Gidaris, Praveer Singh, and Nikos Komodakis.
\newblock Unsupervised representation learning by predicting image rotations.
\newblock In {\em International Conference on Learning Representations}, 2018.

\bibitem{goodfellow2016deep}
Ian Goodfellow, Yoshua Bengio, and Aaron Courville.
\newblock {\em Deep learning}.
\newblock MIT press, 2016.

\bibitem{grill2020bootstrap}
Jean-Bastien Grill, Florian Strub, Florent Altch\'{e}, Corentin Tallec, Pierre Richemond, Elena Buchatskaya, Carl Doersch, Bernardo Avila~Pires, Zhaohan Guo, Mohammad Gheshlaghi~Azar, Bilal Piot, koray kavukcuoglu, Remi Munos, and Michal Valko.
\newblock Bootstrap your own latent - a new approach to self-supervised learning.
\newblock In H.~Larochelle, M.~Ranzato, R.~Hadsell, M.F. Balcan, and H.~Lin, editors, {\em Advances in Neural Information Processing Systems}, volume~33, pages 21271--21284. Curran Associates, Inc., 2020.

\bibitem{ha2016hypernetworks}
David Ha, Andrew~M. Dai, and Quoc~V. Le.
\newblock Hypernetworks.
\newblock In {\em International Conference on Learning Representations}, 2017.

\bibitem{he2022mae}
Kaiming He, Xinlei Chen, Saining Xie, Yanghao Li, Piotr Doll\'ar, and Ross Girshick.
\newblock Masked autoencoders are scalable vision learners.
\newblock In {\em Proceedings of the IEEE/CVF Conference on Computer Vision and Pattern Recognition (CVPR)}, pages 16000--16009, June 2022.

\bibitem{he2020momentum}
Kaiming He, Haoqi Fan, Yuxin Wu, Saining Xie, and Ross Girshick.
\newblock Momentum contrast for unsupervised visual representation learning.
\newblock In {\em Proceedings of the IEEE/CVF Conference on Computer Vision and Pattern Recognition (CVPR)}, June 2020.

\bibitem{he2015deep}
Kaiming He, Xiangyu Zhang, Shaoqing Ren, and Jian Sun.
\newblock Deep residual learning for image recognition.
\newblock In {\em Proceedings of the IEEE Conference on Computer Vision and Pattern Recognition (CVPR)}, June 2016.

\bibitem{hendrycks2019benchmarking}
Dan Hendrycks and Thomas Dietterich.
\newblock Benchmarking neural network robustness to common corruptions and perturbations.
\newblock In {\em International Conference on Learning Representations}, 2019.

\bibitem{kim2022did}
Tae~Soo Kim, Geonwoon Jang, Sanghyup Lee, and Thijs Kooi.
\newblock Did you get what you paid for? rethinking annotation cost of deep learning based computer aided detection in chest radiographs.
\newblock In Linwei Wang, Qi~Dou, P.~Thomas Fletcher, Stefanie Speidel, and Shuo Li, editors, {\em Medical Image Computing and Computer Assisted Intervention -- MICCAI 2022}, pages 261--270, Cham, 2022. Springer Nature Switzerland.

\bibitem{kornblith2019better}
Simon Kornblith, Jonathon Shlens, and Quoc~V. Le.
\newblock Do better imagenet models transfer better?
\newblock In {\em Proceedings of the IEEE/CVF Conference on Computer Vision and Pattern Recognition (CVPR)}, June 2019.

\bibitem{krause2013cars}
Jonathan Krause, Michael Stark, Jia Deng, and Li~Fei-Fei.
\newblock 3d object representations for fine-grained categorization.
\newblock In {\em 4th International IEEE Workshop on 3D Representation and Recognition (3dRR-13)}, Sydney, Australia, 2013.

\bibitem{krizhevsky2009learning}
Alex Krizhevsky.
\newblock Learning multiple layers of features from tiny images.
\newblock 2009.

\bibitem{kumar2022biosignals}
Amit~Krishan Kumar, M.~Ritam, Lina Han, Shuli Guo, and Rohitash Chandra.
\newblock Deep learning for predicting respiratory rate from biosignals.
\newblock {\em Computers in Biology and Medicine}, 144:105338, 2022.

\bibitem{lee2021improving}
Hankook Lee, Kibok Lee, Kimin Lee, Honglak Lee, and Jinwoo Shin.
\newblock Improving transferability of representations via augmentation-aware self-supervision.
\newblock In M.~Ranzato, A.~Beygelzimer, Y.~Dauphin, P.S. Liang, and J.~Wortman Vaughan, editors, {\em Advances in Neural Information Processing Systems}, volume~34, pages 17710--17722. Curran Associates, Inc., 2021.

\bibitem{tsung2014cocodataset}
Tsung{-}Yi Lin, Michael Maire, Serge~J. Belongie, Lubomir~D. Bourdev, Ross~B. Girshick, James Hays, Pietro Perona, Deva Ramanan, Piotr Doll{'{a} }r, and C.~Lawrence Zitnick.
\newblock Microsoft {COCO:} common objects in context.
\newblock {\em CoRR}, abs/1405.0312, 2014.

\bibitem{maji2013finegrained}
Subhransu Maji, Esa Rahtu, Juho Kannala, Matthew~B. Blaschko, and Andrea Vedaldi.
\newblock Fine-grained visual classification of aircraft.
\newblock {\em CoRR}, abs/1306.5151, 2013.

\bibitem{mialon2022variance}
Gr{\'e}goire Mialon, Randall Balestriero, and Yann LeCun.
\newblock Variance covariance regularization enforces pairwise independence in self-supervised representations, 2023.

\bibitem{mnih2015human}
Volodymyr Mnih, Koray Kavukcuoglu, David Silver, Andrei~A Rusu, Joel Veness, Marc~G Bellemare, Alex Graves, Martin Riedmiller, Andreas~K Fidjeland, Georg Ostrovski, et~al.
\newblock Human-level control through deep reinforcement learning.
\newblock {\em nature}, 518(7540):529--533, 2015.

\bibitem{nilsback2008flowers}
Maria-Elena Nilsback and Andrew Zisserman.
\newblock Automated flower classification over a large number of classes.
\newblock In {\em Indian Conference on Computer Vision, Graphics and Image Processing}, Dec 2008.

\bibitem{noroozi2017unsupervised}
Mehdi Noroozi and Paolo Favaro.
\newblock Unsupervised learning of visual representations by solving jigsaw puzzles.
\newblock In Bastian Leibe, Jiri Matas, Nicu Sebe, and Max Welling, editors, {\em Computer Vision -- ECCV 2016}, pages 69--84, Cham, 2016. Springer International Publishing.

\bibitem{oquab2023dinov2}
Maxime Oquab, Timothée Darcet, Théo Moutakanni, Huy Vo, Marc Szafraniec, Vasil Khalidov, Pierre Fernandez, Daniel Haziza, Francisco Massa, Alaaeldin El-Nouby, Mahmoud Assran, Nicolas Ballas, Wojciech Galuba, Russell Howes, Po-Yao Huang, Shang-Wen Li, Ishan Misra, Michael Rabbat, Vasu Sharma, Gabriel Synnaeve, Hu~Xu, Hervé Jegou, Julien Mairal, Patrick Labatut, Armand Joulin, and Piotr Bojanowski.
\newblock Dinov2: Learning robust visual features without supervision.
\newblock {\em ArXiv}, abs/2304.07193, 2023.

\bibitem{ozbulak2023know}
Utku Ozbulak, Hyun~Jung Lee, Beril Boga, Esla~Timothy Anzaku, Ho~min Park, Arnout~Van Messem, Wesley~De Neve, and Joris Vankerschaver.
\newblock Know your self-supervised learning: A survey on image-based generative and discriminative training.
\newblock {\em Transactions on Machine Learning Research}, 2023.
\newblock Survey Certification.

\bibitem{parkhi2012pets}
Omkar~M. Parkhi, Andrea Vedaldi, Andrew Zisserman, and C.~V. Jawahar.
\newblock Cats and dogs.
\newblock In {\em IEEE Conference on Computer Vision and Pattern Recognition}, 2012.

\bibitem{paszke2019pytorch}
Adam Paszke, Sam Gross, Francisco Massa, Adam Lerer, James Bradbury, Gregory Chanan, Trevor Killeen, Zeming Lin, Natalia Gimelshein, Luca Antiga, Alban Desmaison, Andreas Kopf, Edward Yang, Zachary DeVito, Martin Raison, Alykhan Tejani, Sasank Chilamkurthy, Benoit Steiner, Lu~Fang, Junjie Bai, and Soumith Chintala.
\newblock Pytorch: An imperative style, high-performance deep learning library.
\newblock In {\em Advances in Neural Information Processing Systems 32}, pages 8024--8035. Curran Associates, Inc., 2019.

\bibitem{quattoni2009mit67}
Ariadna Quattoni and Antonio Torralba.
\newblock Recognizing indoor scenes.
\newblock In {\em 2009 IEEE Conference on Computer Vision and Pattern Recognition}, pages 413--420, 2009.

\bibitem{raghu2021metalearning}
Aniruddh Raghu, Jonathan Lorraine, Simon Kornblith, Matthew McDermott, and David~K Duvenaud.
\newblock Meta-learning to improve pre-training.
\newblock In M.~Ranzato, A.~Beygelzimer, Y.~Dauphin, P.S. Liang, and J.~Wortman Vaughan, editors, {\em Advances in Neural Information Processing Systems}, volume~34, pages 23231--23244. Curran Associates, Inc., 2021.

\bibitem{ren2015faster}
Shaoqing Ren, Kaiming He, Ross Girshick, and Jian Sun.
\newblock Faster r-cnn: Towards real-time object detection with region proposal networks.
\newblock In C.~Cortes, N.~Lawrence, D.~Lee, M.~Sugiyama, and R.~Garnett, editors, {\em Advances in Neural Information Processing Systems}, volume~28. Curran Associates, Inc., 2015.

\bibitem{robinson2021contrastive}
Joshua Robinson, Li~Sun, Ke~Yu, Kayhan Batmanghelich, Stefanie Jegelka, and Suvrit Sra.
\newblock Can contrastive learning avoid shortcut solutions?
\newblock In M.~Ranzato, A.~Beygelzimer, Y.~Dauphin, P.S. Liang, and J.~Wortman Vaughan, editors, {\em Advances in Neural Information Processing Systems}, volume~34, pages 4974--4986. Curran Associates, Inc., 2021.

\bibitem{russakovsky2015imagenet}
Olga Russakovsky, Jia Deng, Hao Su, Jonathan Krause, Sanjeev Satheesh, Sean Ma, Zhiheng Huang, Andrej Karpathy, Aditya Khosla, Michael Bernstein, Alexander Berg, and Li~Fei-Fei.
\newblock Imagenet large scale visual recognition challenge.
\newblock {\em International Journal of Computer Vision}, 115, 09 2014.

\bibitem{sagonas2016300w}
C.~Sagonas, E.~Antonakos, Tzimiropoulos G, S.~Zafeiriou, and M.~Pantic.
\newblock 300 faces in-the-wild challenge: Database and results.
\newblock In {\em Image and Vision Computing (IMAVIS), Special Issue on Facial Landmark Localisation}, 2016.

\bibitem{schiappa2022self}
Madeline~C. Schiappa, Yogesh~S. Rawat, and Mubarak Shah.
\newblock Self-supervised learning for videos: A survey.
\newblock {\em {ACM} Computing Surveys}, dec 2022.

\bibitem{tian2023designing}
Keyu Tian, Yi~Jiang, qishuai diao, Chen Lin, Liwei Wang, and Zehuan Yuan.
\newblock Designing {BERT} for convolutional networks: Sparse and hierarchical masked modeling.
\newblock In {\em The Eleventh International Conference on Learning Representations}, 2023.

\bibitem{tian2020makes}
Yonglong Tian, Chen Sun, Ben Poole, Dilip Krishnan, Cordelia Schmid, and Phillip Isola.
\newblock What makes for good views for contrastive learning?
\newblock 33:6827--6839, 2020.

\bibitem{tian2022understanding}
Yuandong Tian.
\newblock Understanding deep contrastive learning via coordinate-wise optimization.
\newblock In Alice~H. Oh, Alekh Agarwal, Danielle Belgrave, and Kyunghyun Cho, editors, {\em Advances in Neural Information Processing Systems}, 2022.

\bibitem{oord2019representation}
A{\"{a}}ron van~den Oord, Yazhe Li, and Oriol Vinyals.
\newblock Representation learning with contrastive predictive coding.
\newblock {\em CoRR}, abs/1807.03748, 2018.

\bibitem{wagner2022importance}
Diane Wagner, Fabio Ferreira, Danny Stoll, Robin~Tibor Schirrmeister, Samuel M{\"u}ller, and Frank Hutter.
\newblock On the importance of hyperparameters and data augmentation for self-supervised learning.
\newblock In {\em First Workshop on Pre-training: Perspectives, Pitfalls, and Paths Forward at ICML 2022}, 2022.

\bibitem{wah2011cub}
C.~Wah, S.~Branson, P.~Welinder, P.~Perona, and S.~Belongie.
\newblock {The Caltech-UCSD Birds-200-2011 Dataset}.
\newblock Technical Report CNS-TR-2011-001, California Institute of Technology, 2011.

\bibitem{wickstrom2022mixing}
Kristoffer Wickstr{\o}m, Michael Kampffmeyer, Karl~{\O}yvind Mikalsen, and Robert Jenssen.
\newblock Mixing up contrastive learning: Self-supervised representation learning for time series.
\newblock {\em Pattern Recognition Letters}, 155:54--61, mar 2022.

\bibitem{xiao2010sun}
J.~{Xiao}, J.~{Hays}, K.~A. {Ehinger}, A.~{Oliva}, and A.~{Torralba}.
\newblock Sun database: Large-scale scene recognition from abbey to zoo.
\newblock In {\em 2010 IEEE Computer Society Conference on Computer Vision and Pattern Recognition}, pages 3485--3492, June 2010.

\bibitem{xiao2020whatshouldnotbecontrastive}
Tete Xiao, Xiaolong Wang, Alexei~A Efros, and Trevor Darrell.
\newblock What should not be contrastive in contrastive learning.
\newblock In {\em International Conference on Learning Representations}, 2021.

\bibitem{xie2022whatshouldbeequivariant}
Yuyang Xie, Jianhong Wen, Kin~Wai Lau, Yasar Abbas~Ur Rehman, and Jiajun Shen.
\newblock What should be equivariant in self-supervised learning.
\newblock In {\em Proceedings of the IEEE/CVF Conference on Computer Vision and Pattern Recognition (CVPR) Workshops}, pages 4111--4120, June 2022.

\bibitem{xie2022simmim}
Zhenda Xie, Zheng Zhang, Yue Cao, Yutong Lin, Jianmin Bao, Zhuliang Yao, Qi~Dai, and Han Hu.
\newblock Simmim: A simple framework for masked image modeling.
\newblock In {\em Proceedings of the IEEE/CVF Conference on Computer Vision and Pattern Recognition (CVPR)}, pages 9653--9663, June 2022.

\bibitem{yosinski2014transferable}
Jason Yosinski, Jeff Clune, Yoshua Bengio, and Hod Lipson.
\newblock How transferable are features in deep neural networks?
\newblock In Z.~Ghahramani, M.~Welling, C.~Cortes, N.~Lawrence, and K.Q. Weinberger, editors, {\em Advances in Neural Information Processing Systems}, volume~27. Curran Associates, Inc., 2014.

\bibitem{zbontar2021barlow}
Jure Zbontar, Li~Jing, Ishan Misra, Yann LeCun, and Stephane Deny.
\newblock Barlow twins: Self-supervised learning via redundancy reduction.
\newblock In Marina Meila and Tong Zhang, editors, {\em Proceedings of the 38th International Conference on Machine Learning}, volume 139 of {\em Proceedings of Machine Learning Research}, pages 12310--12320. PMLR, 18--24 Jul 2021.

\bibitem{zhang2019aet}
Liheng Zhang, Guo-Jun Qi, Liqiang Wang, and Jiebo Luo.
\newblock Aet vs. aed: Unsupervised representation learning by auto-encoding transformations rather than data.
\newblock In {\em Proceedings of the IEEE/CVF Conference on Computer Vision and Pattern Recognition (CVPR)}, June 2019.

\bibitem{zhang2016colorful}
Richard Zhang, Phillip Isola, and Alexei~A. Efros.
\newblock Colorful image colorization.
\newblock In Bastian Leibe, Jiri Matas, Nicu Sebe, and Max Welling, editors, {\em Computer Vision -- ECCV 2016}, pages 649--666, Cham, 2016. Springer International Publishing.

\bibitem{zini2023planckian}
Simone Zini, Alex Gomez-Villa, Marco Buzzelli, Bart{\l}omiej Twardowski, Andrew~D. Bagdanov, and Joost van~de weijer.
\newblock Planckian jitter: countering the color-crippling effects of color jitter on self-supervised training.
\newblock In {\em The Eleventh International Conference on Learning Representations}, 2023.

\end{thebibliography}
}

\appendix

\section{Pretraining and evaluation details}
\label{sec:app:details}

In this section, we describe the details of Self-Supervised pretraining and the evaluation methodology used in our experiments.

\subsection{Pretraining}
\label{sec:app:train}

\paragraph{Datasets}
We use ImageNet-100, a 100-class subset of ImageNet \cite{russakovsky2015imagenet,tian2020makes}, to pretrain the standard ResNet-50 \cite{he2015deep} architecture of self-supervised methods: MoCo-v2 \cite{chen2020improved}, SimCLR \cite{chen2020simple}, Barlow Twins \cite{zbontar2021barlow}, 
as well as for common in the literature on augmentation-aware self-supervised learning \cite{tian2020makes,xiao2020whatshouldnotbecontrastive,lee2021improving,chavhan2023amortised}. For MoCo-v3 \cite{chen2021empirical}, we pretrain the ViT-Small \cite{dosovitskiy2021image} model on the full ImageNet dataset \cite{russakovsky2015imagenet}.

\paragraph{Hyperparameters}

We follow the pretraining procedures from corresponding papers, described in \cite{lee2021improving} for MoCo-v2
, \cite{chavhan2023amortised} for SimCLR, \cite{zbontar2021barlow} for Barlow Twins, and \cite{chen2021empirical} for MoCo-v3.  Synchronized batch normalization is employed for distributed training \cite{chen2021exploring}.
In Table \ref{tab:hyperparams}, we present the training hyperparameters which are not related specifically to \our{}, but rather joint-embedding approaches in general~\cite{he2020momentum,chen2020improved,chen2020simple,zbontar2021barlow,chen2021exploring,chen2021empirical}.

\begin{table*}[h]

    \centering

    \small
      \setlength\tabcolsep{3pt}
    \caption{Hyperparameters of self-supervised methods used with \our{}}
      \resizebox{\linewidth}{!}{
    \begin{tabular}{cccccccccccc}
         \toprule
         \textbf{SSL method} & \textbf{Architecture} & \textbf{Number of } &  \textbf{Batch}  & \textbf{Weight} & \multicolumn{2}{c}{\textbf{Learning rate}} & \textbf{Training} \\
                            &                        & \textbf{epochs }       &  \textbf{size} & \textbf{decay} & \textit{Base} & \textit{Schedule}  & \textbf{time}\\
         \midrule
         MoCo-v2 \cite{chen2020improved} & ResNet-50 & 500 & 256 & $10^{-4}$ & 0.03 & Cosine decay & 34h \\
         SimCLR \cite{chen2020simple} & ResNet-50 & 300 & 1024 & $10^{-4}$ & 0.05 & Cosine decay & 10h \\
         Barlow Twins \cite{zbontar2021barlow} & ResNet-50 & 500 & 256 & $10^{-4}$ & 0.05 & Cosine decay with warmup & 36h \\
          MoCo-v3 \cite{chen2021empirical} & ViT-small & 300 & 1024 & 0.1 & $1.5 \cdot 10^{-4}$ & Cosine decay with warmup & 23h \\

         \bottomrule
    \end{tabular}
    }
      \resizebox{\linewidth}{!}{
    \begin{tabular}{cccccccccccc}
         \toprule
         \textbf{SSL method} & \multicolumn{4}{c}{\textbf{Projector}}  & \multicolumn{4}{c}{\textbf{Predictor}}  \\
                            &  \textit{Depth} & \textit{Hidden size} & \textit{Out size} &\textit{Final BatchNorm}  &  \textit{Depth} & \textit{Hidden size} & \textit{Out size} &\textit{Final BatchNorm}\\
         \midrule

         MoCo-v2  & 2 & 2048 & 128 & No  & \multicolumn{4}{c}{No} \\
         SimCLR &  2 & 2048 & 128 & No  & \multicolumn{4}{c}{No} \\
         Barlow Twins  &  3 & 8192 & 8192 & Yes, without affine transform  & \multicolumn{4}{c}{No} \\
        MoCo-v3 &  3 & 4096 & 256 & Yes, without affine transform & 2 & 4096 & 256 & Yes, without affine transform  \\

         \bottomrule
    \end{tabular}
    }
    \label{tab:hyperparams}
\end{table*}

\paragraph{Augmentations}

For self-supervised pretraining, we use a set of augmentations adopted commonly in the literature \cite{he2020momentum,chen2020simple,chen2020improved,chen2021exploring,zbontar2021barlow,lee2021improving}. We denote them below:

\begin{itemize}
    \item \textbf{random cropping}  -- We sample the cropping scale randomly from [0.2, 1.0]. Afterward, we resize the cropped images to the size of $224 \times 224$.
    \item \textbf{color jittering} -- We apply this operation with a probability of $0.8$. We sample the intensities of brightness, contrast, saturation, and hue and their maximal values are $0.4$, $0.4$, $0.4$, and $0.1$, respectively.
    \item \textbf{Gaussian blurring} -- We apply this operation with a probability of $0.5$. We sample the standard deviation from $[0.1, 2.0]$ and set the kernel size to $23 \times 23$.
    \item \textbf{random horizontal flipping} -- We apply this operation with a probability of $0.5$.
    \item \textbf{random grayscaling} -- We apply this operation with a probability of $0.2$.
\end{itemize}

\subsection{Evaluation}
\label{sec:app:eval}

\paragraph{Linear evaluation}

The linear evaluation (linear probing) protocol used throughout this work follows  \cite{chen2020simple,grill2020bootstrap,kornblith2019better,lee2021improving}. Namely, we center-crop and resize the images from the downstream dataset to the size of $224 \times 224$, pass them through the pretrained feature extractor, and obtain the embeddings from the final feature extractor stage. The only exception from this is the CUB dataset \cite{wah2011cub}, where, following \cite{lee2021improving}, for the training images besides the center crop of the image, we also crop the image at its corners and do the same for the horizontal flip of the image (this is known as TenCrop operation\footnote{\url{https://pytorch.org/vision/main/generated/torchvision.transforms.TenCrop.html}}). 
Having gathered the image features, we minimize the $l_2$-regularized cross-entropy objective using L-BFGS on the features of the training images. We select the regularization parameter from between $[10^{-6}, 10^5]$ using the validation features. Finally, we train the linear classifier on training and validation features with the selected $l_2$ parameter and report the final performance metric (see Table \ref{tab:eval_datasets}) on the test dataset. We set the maximum number of iterations in L-BFGS as 5000 and use the model trained on training data as initialization for training the final model. 

We note that the above linear evaluation procedure is effectively equivalent to training the final layer of a network on non-augmented data while keeping the remainder of the parameters unchanged.

We list the datasets and evaluation metrics used during linear evaluation in Table~\ref{tab:eval_datasets}.

\begin{table*}[h]
    \centering
    \caption{Datasets and respective evaluation metrics used for linear evaluation of \our{}.}
          \resizebox{\linewidth}{!}{
    \begin{tabular}{ccc}
        \toprule
         \textbf{Downstream task} & \textbf{Dataset} &\textbf{Evaluation metric} \\
         \midrule
         C10 & CIFAR10 \cite{krizhevsky2009learning} & Top-1 accuracy \\
         C100 & CIFAR100 \cite{krizhevsky2009learning} &  Top-1 accuracy \\
         Food & Food101 \cite{bossard2014food101} & Top-1 accuracy \\
         MIT & MIT67 \cite{quattoni2009mit67} & Top-1 accuracy \\
         Pets & Oxford-IIIT Pets \cite{parkhi2012pets} & Mean per-class accuracy \\
         Flowers & Oxford Flowers-102 \cite{nilsback2008flowers} & Mean per-class accuracy \\
         Caltech & Caltech101 \cite{li2006caltech} & Mean per-class accuracy \\
         Cars & Stanford Cars \cite{krause2013cars} & Top-1 accuracy \\
         FGVCA & FGVC-Aircraft \cite{maji2013finegrained} &  Mean Per-class accuracy \\
         DTD & Describable Textures  (split 1) \cite{cimpoi14describing} & Top-1 accuracy \\
         SUN & SUN397 (split 1) \cite{xiao2010sun} & Top-1 accuracy \\
         CUB & Caltech-UCSD Birds \cite{wah2011cub} & Top-1 accuracy \\
         300W & 300 Faces In-the-Wild \cite{sagonas2016300w} & R$^2$ \\
         \bottomrule
    \end{tabular}
    }
    \label{tab:eval_datasets}
\end{table*}

\paragraph{Object detection}

We closely follow the evaluation protocol of \cite{he2020momentum}. We train the Faster-RCNN \cite{ren2015faster} model with the pretrained backbone. Contrary to  \cite{he2020momentum}, we do not tune the backbone parameters, in order to better observe the effect of different pretraining methods. We report the Average Precision \cite{tsung2014cocodataset} measured on the VOC \texttt{test2007} set \cite{everingham2007pascalvoc}.

\paragraph{Sensitivity to augmentations}

We consider image pairs, where one image is the (center-cropped) original and the second one is augmented by the given augmentation. For each image pair, we extract image features at four stages of the pretrained ResNet-50 backbone \cite{he2015deep}, as well as the final representation of the projector network. We next calculate cosine similarities between the features of augmented and non-augmented images in the given mini-batch (of size 256). We report the value of the InfoNCE loss~\cite{oord2019representation} calculated on such similarity matrices.

\paragraph{Dependency of \our{} projector on conditioning} Similarly to the above experiment, we compare the projector features of augmented and non-augmented image pairs. When computing the features of the augmented image, we supply the projector with the augmentation embedding computed from augmentation parameters corresponding to either this image (true augmentation information) or another, randomly chosen image from the same mini-batch (random augmentation information). We then compute cosine similarities between the original image features and features of the augmented image computed with true/random augmentation information.

\subsection{Implementation details}
\label{sec:impl}

We implement \our{} in the PyTorch framework \cite{paszke2019pytorch}, building upon the codebase of \cite{lee2021improving}. Our code is available at \href{https://github.com/gmum/CASSLE}{\url{github.com/gmum/CASSLE}}. We train variants of all SSL approaches on 2 NVidia A100 GPUs, except for SimCLR and MoCo-v3 which use larger batch sizes and therefore require 8 such GPUs.

\section{Additional analysis of \our{}}
\label{sec:exp:analysis}

In this section, we highlight several additional features of \our{}, including its effect on the pretraining loss minimization, generalization to unseen augmentations, and robustness to perturbed images. 
We also compare \our{} with the masked image modeling family of SSL methods~\cite{tian2023designing}, and demonstrate its applicability to joint-embedding SSL approaches that do not utilize the projector in their architecture~\cite{he2020momentum}.

\subsection{Analysis of the self-supervised learning procedure}
\label{sec:exp:contrastive_loss}

\begin{figure}[h]
    \centering
    \includegraphics[width=\textwidth]{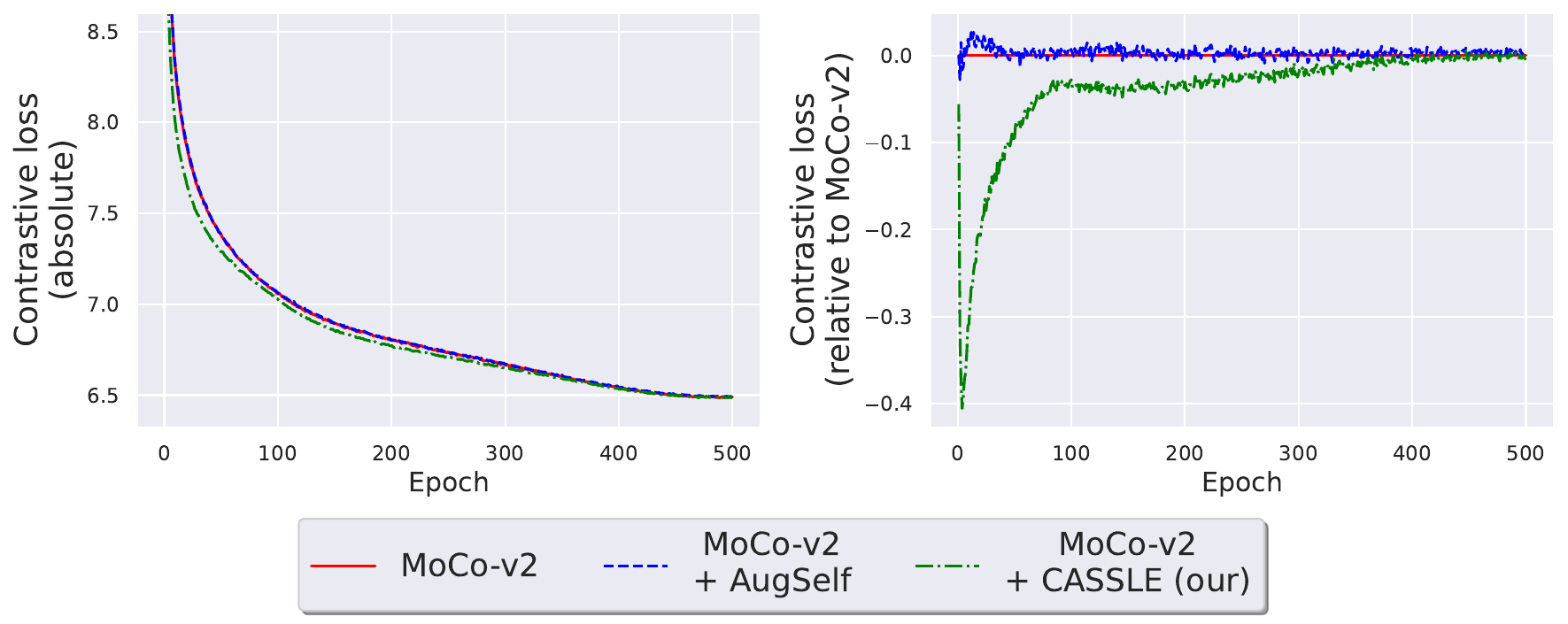}
    \caption{Absolute (left) and relative to Baseline (right) values of contrastive losses of Baseline, AugSelf \cite{lee2021improving}, and \our{} MoCo-v2 variants, measured during pretraining. \our{} minimizes the contrastive objective faster than  Baseline and AugSelf, in particular early in the training procedure.
    }
    \label{fig:exp_losses}
    \vspace{-0.2cm}
\end{figure}

We compare the training of MoCo-v2 \cite{he2020momentum,chen2020improved} with and without \our{} or AugSelf~\cite{lee2021improving} extensions, and plot the contrastive loss values measured throughout training in the left part of Figure \ref{fig:exp_losses}, and on the right, the values of losses relative to the vanilla MoCo-v2. 
\our{} minimizes the contrastive objective faster than the other two variants, in particular early in the training procedure. This suggests that augmentation information provides helpful conditioning for a model not yet fully trained to align augmented image pairs and thus, \our{} learns to depend on this information. 

{
\subsection{Modulating the augmentation-awareness}
As seen in Section~\ref{sec:exp:cosine}, \our{} increases the augmentation-awareness of feature extractor representations. This raises a question -- \emph{can we influence the level of augmentation-awareness?} 

In \our{}, the task of augmentation-awareness is not enforced by any specific objective function. We consider this as an advantage of \our{} -- the model can learn to use the augmentation information to a degree that is useful for solving the invariance task. On the other hand, in AugSelf~\cite{lee2021improving}, there is a need to balance the invariance and sensitivity objectives with a hyperparameter.
In \our{}, we can modulate the emergent augmentation-awareness by modifying the expressiveness of the augmentation encoder network. Thus, if the amount of available augmentation information during training is reduced, the model should learn not to rely on it and become more augmentation-invariant. To verify this, we train several variants of MoCo-v2+\our{} with increased and compressed \guidingnet{} output sizes. Next, similarly to the experiment described in Section~\ref{sec:exp:cosine}, we compare the InfoNCE loss of matching the feature extractor output representations of augmented image pairs. We show the results for different augmentation types in Figure~\ref{fig:exp_invariance_outsize}. Models trained with a less expressive \guidingnet{} are also more augmentation-invariant, as evidenced by lower InfoNCE values. This can be attributed to the fact that augmentation information compressed low-dimensional embedding may be less informative for solving the augmentation invariance task.

\begin{figure}
    \centering
    \includegraphics[width=\textwidth]{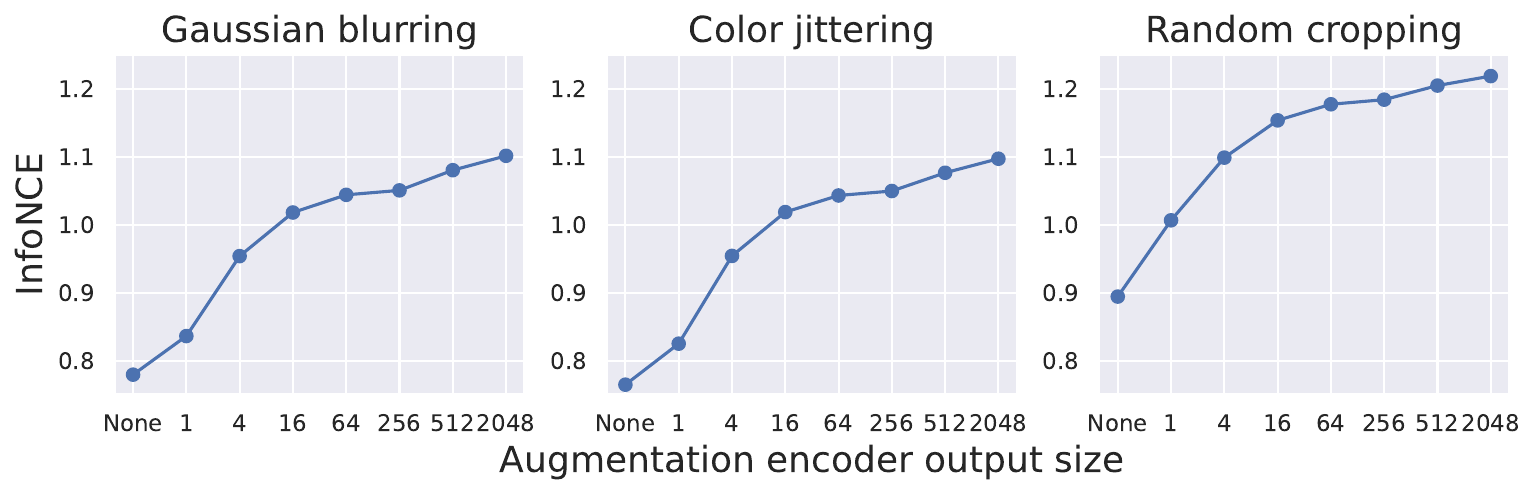}
    \caption{{A comparison of InfoNCE loss measured on different kinds of augmentations at the output of the feature extractor. We compare variants of MoCo-v2+\our{} with different \guidingnet{} output sizes, as well as the vanilla MoCo-v2 (denoted as \emph{None}). 
    \our{} with increasing \guidingnet{} sizes yield higher InfoNCE values, which suggests that this hyperparameter influences the augmentation sensitivity of \our{}.}
    }
    \label{fig:exp_invariance_outsize}
\end{figure}

}

\subsection{Generalization to unseen augmentations} 
\label{sec:exp:rotation}

\begin{table}[h]
  \caption{Linear evaluation on the task of predicting image rotation ($0$/$90$/$180$/$270$ degrees)}
  \small
  \label{tab:rotation}
  \centering
  \setlength\tabcolsep{2.5pt}
  \resizebox{\linewidth}{!}{
 \begin{tabular}{lccccccccccc} 
    \toprule
    \textbf{Method} & C10 & C100 & Food & Pets & MIT & Flowers & Caltech & Cars & FGVCA & STL10 & SUN \\
    \toprule
    & \multicolumn{10}{c}{\textit{SimCLR} \cite{chen2020simple}} \\
    \midrule
    
    Vanilla & 67.93 & 56.88 & 68.76 & 65.07 & \textbf{70.40} & 24.15 & 56.81 & 87.12 & 98.02 & \textbf{73.90} & 67.94 \\
    AugSelf \cite{lee2021improving} & 75.63 & 63.08 & 75.01 & 57.99 & 16.54 & \textbf{33.99} & 39.03 & 76.11 & 96.64 & 67.88 & 76.60 \\
    \textbf{\our{}} & \textbf{77.07} & \textbf{70.20} & \textbf{79.55} & \textbf{70.97} & 68.77 & 33.08 & \textbf{64.97} & \textbf{95.14} & \textbf{99.46} & 72.69 & \textbf{78.16} \\

    \midrule
    
    & \multicolumn{10}{c}{\textit{MoCo-v2} \cite{he2020momentum,chen2020improved}} \\
    \midrule
    Vanilla & 67.96 & 56.96 & 75.70 & 71.27 & 67.13 & \textbf{58.30} & \textbf{87.33} & 58.35 & 93.20 & \textbf{95.44} & 68.33  \\
      AugSelf \cite{lee2021improving} & \textbf{74.57}&  65.87 & 73.03 & \textbf{77.01} & \textbf{79.75} & 52.81 & 58.53 & \textbf{93.07} & 98.26 & 77.35 & \textbf{83.66} \\
    \textbf{\our{}} & 73.21 & \textbf{69.91} & \textbf{77.31} & 76.04 & 76.40 & 45.11 & 61.06 & 90.49 & \textbf{98.56} & 63.65 & 76.01 \\
    
    \midrule

    & \multicolumn{10}{c}{\textit{Barlow Twins} \cite{zbontar2021barlow}} \\
    \midrule
    
    Vanilla & 73.67 & 64.87 & 72.85 &  \textbf{83.02} & 	70.97	& 42.56 &	\textbf{57.71}	& 83.68	& 97.93	& 65.70 & 	76.87 \\
      AugSelf \cite{lee2021improving} & 72.79	& \textbf{65.91} & 	74.93	& 77.76	& 50.30	& 30.93	& 44.31	& 85.85	& 98.35	& \textbf{69.68}	& \textbf{77.19} \\
    \textbf{\our{}} &  \textbf{74.97} & 	65.15 & \textbf{76.55} & 77.05 & \textbf{86.49} & \textbf{43.16} & 53.60 & 	\textbf{92.92} & 	\textbf{99.16} & 	55.24 &	74.97 \\
    
    \bottomrule
  \end{tabular}
  } 
  \vspace{-0.5cm}
\end{table}

To understand whether \our{} generalizes to types of augmentation that were not used during pretraining, we inspect its performance in the task of prediction of the applied augmentation. We train a linear classifier on top of the pretrained model to predict whether the image was rotated by $0$, $90$, $180$, or $270$ degrees. We formulate the problem as classification due to its cyclic nature and test the model on the same datasets as in Section \ref{sec:exp:downstream}. We present the results of vanilla, AugSelf~\cite{lee2021improving} and \our{} variants of self-supervised methods in Table \ref{tab:rotation}. Apart from a few exceptions, \our{} and AugSelf extensions allow in general for better rotation prediction than the vanilla SSL methods. Moreover, in the case of SimCLR and Barlow Twins, the \our{} representation predicts the rotations the most accurately on a vast majority of datasets. This occurs despite the fact, that neither of the methods was trained using rotated images and thus, never explicitly learned the \emph{concept} of rotation. 
This suggests that \our{} learns representations that are sensitive to a broader set of perturbations than those whose information had been used during pretraining.

\subsection{Robustness under perturbations}

We next verify the influence of increased sensitivity to augmentations on the robustness to perturbations of models pretrained with MoCo-v2~\cite{he2020momentum,chen2020improved} and SimCLR~\cite{chen2020simple}. Following the experimental setup of \cite{lee2021improving}, we train the pretrained networks for classification of ImageNet-100 and evaluate them on weather-corrupted images (fog, frost, snow, and brightness)~\cite{hendrycks2019benchmarking} from the validation set. We report the results in Table \ref{tab:app:perturbation}.
We find that the network pretrained with MoCo-v2+\our{} achieves the best results when dealing with images perturbed by brightness and snow, whereas vanilla MoCo-v2 performs best on images perturbed by fog and frost. When it comes to SimCLR, except for the images perturbed by frost, \our{} achieves the best performance.

\begin{table}[h]
\centering
  \caption{Evaluation of variants of MoCo-v2 and SimCLR on perturbed ImageNet-100 images.}
      \label{tab:app:perturbation}
  \resizebox{\linewidth}{!}{
         \begin{tabular}{lcccc|ccccccccc} 
    \toprule
    
    \textbf{Method} & \multicolumn{4}{c|}{\it MoCo-v2~\cite{he2020momentum,chen2020improved}} & \multicolumn{4}{c}{\it SimCLR~\cite{chen2020simple}} \\
    \textbf{Perturbation} & Brightness & Frost & Fog & Snow & Brightness & Frost & Fog & Snow\\
    \midrule
    \textbf{Vanilla} & 85.30 & \textbf{53.70} & \textbf{56.92} & 31.78 & 85.74 & \textbf{50.66} & 53.94 & 33.78 \\
    \textbf{AugSelf} & 83.64 & 51.98 & 53.08 & 33.80 & 85.84 & 50.34 & 53.60 & 32.98 \\
    \textbf{\our{}} & \textbf{86.10} & 50.54 & 54.22 & \textbf{34.66} &  \textbf{86.38} & 48.62 & \textbf{59.04} & \textbf{35.30} \\
    
    \bottomrule

    \end{tabular}
    }
    \vspace{-0.5cm}
\end{table}

\subsection{Comparison with Masked Image Modeling}

Recently, Masked Image Modeling (MIM) methods have emerged as a new family of approaches to Self-supervised Learning~\cite{xie2022simmim,he2022mae,tian2023designing}. Contrary to Joint-Embedding methods, MIM-based methods rely on missing data imputation as their pretext task. Thus, a natural question arises, whether MIM-based methods are not a better tool to address the cases where augmentation-invariance is expected to be problematic?

{To answer this, we compare the vanilla MoCo-v2, and MoCo-v2+\our{} with Sparse masKed modeling (SparK)~\cite{tian2023designing} -- a recently introduced method for Masked Image Modeling for convolutional networks.} For consistency, we use the ResNet-50 backbone with all models and pretrain SparK with hyperparameters suggested by the authors~\cite{tian2023designing}. We compare the methods in terms of augmentation invariance and linear evaluation, analogously to Sections \ref{sec:exp:cosine} and \ref{sec:exp:downstream}, respectively.

Given that SparK does not rely on the augmentation-invariance objective, we do not expect its representations to be invariant to augmentations.
\begin{figure}[h]
    \centering
    \includegraphics[width=\textwidth]{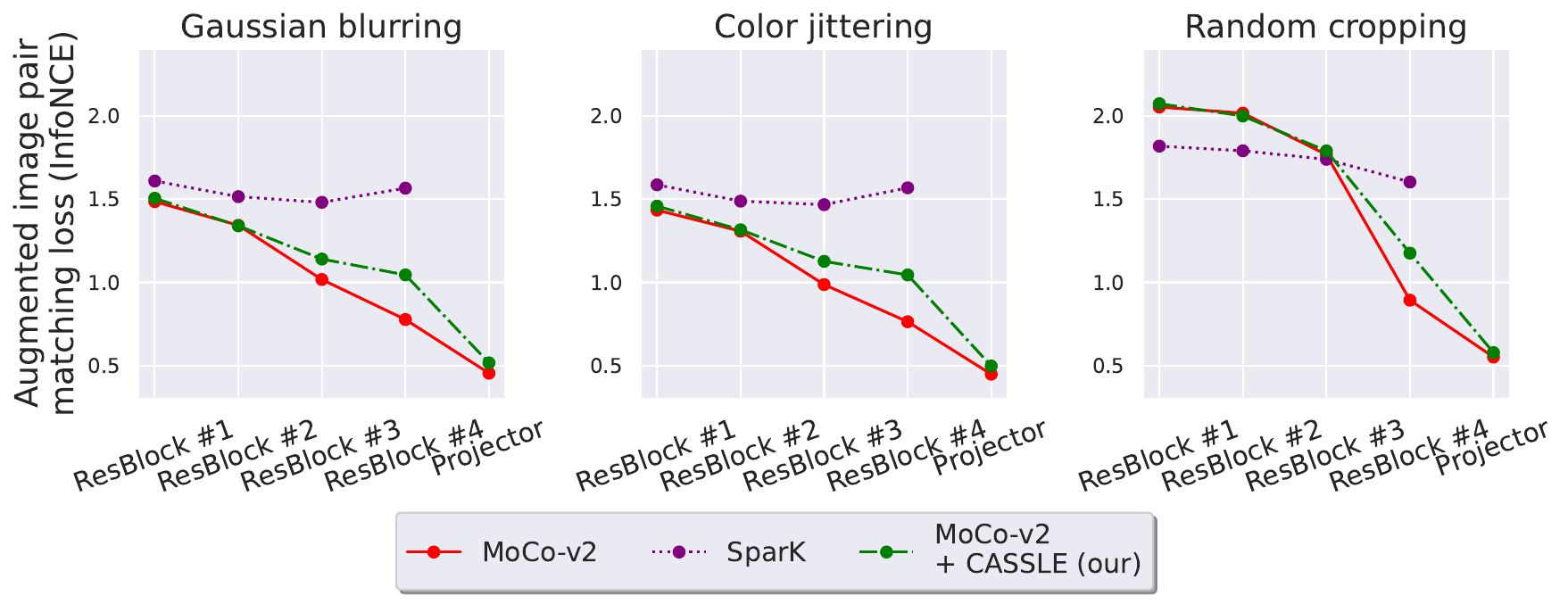}
    \caption{Comparison of augmentation-awareness of vanilla and \our{} variants of MoCo-v2~\cite{he2020momentum,chen2020improved} and SparK~\cite{tian2023designing}. Since SparK is not trained through a Joint-Embedding objective of augmentation-invariance, its representations exhibit higher sensitivity to changes in data caused by augmentations compared to the joint-embedding MoCo-v2. }
    \label{fig:exp_invariance_spark}
\end{figure}

\noindent We confirm this in Figure~\ref{fig:exp_invariance_spark} -- SparK yields high values of the InfoNCE loss of matching the embeddings of augmented image pairs. This indicates that its representations are highly sensitive to features transformed by augmentations. 

However, the linear evaluation results shown in Table~\ref{tab:linear_spark} show that SparK representations perform much worse in the majority of downstream tasks compared to both vanilla MoCo-v2 and MoCo-v2+\our{}.  Indeed, poor linear evaluation accuracy of MIM-based models compared to Joint-Embedding models is their known drawback~\cite{ozbulak2023know,tian2023designing}. 
\begin{table*}[h]
  \caption{Linear evaluation comparison of vanilla and \our{} variants of MoCo-v2~\cite{he2020momentum,chen2020improved} and SparK~\cite{tian2023designing}. All models utilize the ResNet-50 architecture~\cite{he2015deep}. Both versions of MoCo-v2 exhibit better results than SparK, with  MoCo-v2+\our{} performing best on the majority of downstream tasks.}
  \small
  \label{tab:linear_spark}
  \centering
  \setlength\tabcolsep{2.5pt}
  \resizebox{\linewidth}{!}{
 \begin{tabular}{lccccccccccccc} 
    \toprule
    \textbf{Method} & C10 & C100 & Food & MIT & Pets & Flowers & Caltech & Cars & FGVCA & DTD & SUN & CUB & 300W  \\ 
    \toprule
   

    MoCo-v2~\cite{he2020momentum,chen2020improved} & 84.60 & 61.60 & 59.67 & 61.64 & 70.08 & 82.43 & 77.25 & 33.86 & 41.21 & 64.47 & 46.50 & 32.20 & $88.77^\dag$ \\
    \textbf{MoCo-v2+ \our{}} & \textbf{86.32}	& \textbf{65.29}	& \textbf{61.93}		& \textbf{63.86}	& 	\textbf{72.86}	& 	\textbf{86.51}	& 	\textbf{79.63}	& 	\textbf{38.82}	& 	\textbf{42.03}		& \textbf{66.54}		& \textbf{49.25}		& \textbf{36.22}		& 88.93 \\
    SparK \cite{tian2023designing} & 84.39 & 60.52 & 49.20 & 52.84 & 51.55 & 74.28 & 74.89 & 23.72 & 33.69 & 57.82 & 39.12 & 23.09 & \textbf{96.13} \\
    \bottomrule
    
    \end{tabular}
    }
\end{table*}

Joint-Embedding methods, even when extended with \our{}, form more augmentation-invariant representations compared to MIM-based methods. Nevertheless, Joint-Embedding methods offer better transfer learning performance, which can be further boosted by extending them with \our{}.

\subsection{Augmentation-aware conditioning without the projector}

\begin{figure*}[h]
    \centering
    \includegraphics[width=\textwidth]{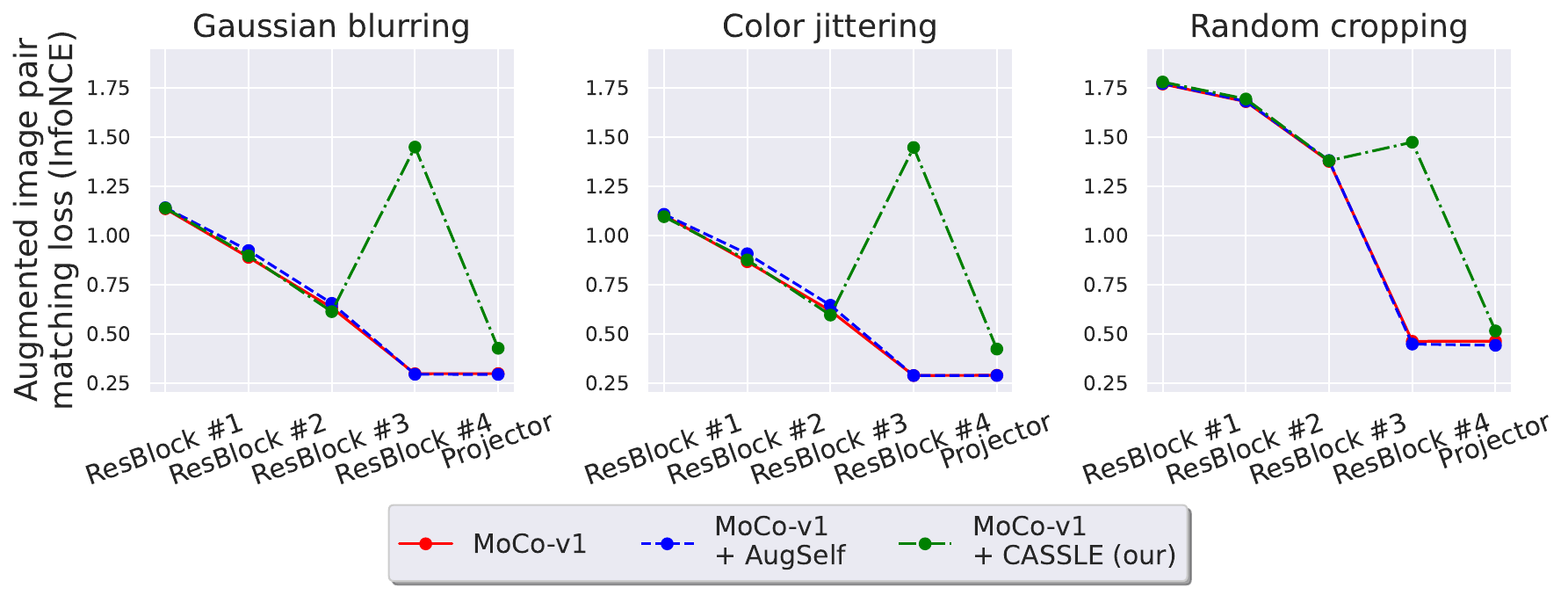}
    \caption{Comparison of augmentation-awareness of vanilla, AugSelf~\cite{lee2021improving}, and \our{} variants of MoCo-v1, which does not contain a projector network. In the case of the vanilla and AugSelf variants, the representation of the fourth ResNet block stage is equivalent to the representation at the projector stage, whereas in \our{}, the projector stage represents the fourth ResNet block stage with added augmentation embeddings.}
    \label{fig:invariance_mocov1}
    \vspace{-0.5cm}
\end{figure*}

Joining image and augmentation embeddings through point-wise addition or multiplication allows us to implement \our{} in Joint-Embedding frameworks that do not utilize the projector in their architecture, such as MoCo-v1~\cite{he2020momentum}. We compare the vanilla, AugSelf, and \our{} (point-wise addition) variants of MoCo-v1 in Table \ref{tab:linear_mocov1}. Surprisingly, \our{} lends a major performance boost to MoCo-v1 despite that it does not utilize a projector network. This suggests that the \our{} augmentation encoder can directly modulate the image embeddings through simple addition. This is further evidenced by Figure~\ref{fig:invariance_mocov1}, where \our{} increases the sensitivity to augmentations of the final stage of network representation by a significant margin. On the other hand, augmentation embeddings produced by \our{} lead to making representations more similar, as visible at the projector stage.

\begin{table*}[h]
  \caption{Linear evaluation of MoCo-v1~\cite{he2020momentum} on downstream classification and regression tasks. \our{} improves the performance of the model by a large margin.}
  \small
  \label{tab:linear_mocov1}
  \centering
  \setlength\tabcolsep{2.5pt}
  \resizebox{\linewidth}{!}{
 \begin{tabular}{lccccccccccccc} 
    \toprule
    \textbf{Method} & C10 & C100 & Food & MIT & Pets & Flowers & Caltech & Cars & FGVCA & DTD & SUN & CUB & 300W  \\ 
    \toprule
   
    MoCo-v1~\cite{he2020momentum} & 58.82 & 28.09 & 25.90 & 31.04 & 47.25 & 33.29	 & 44.41 & 5.00 & 	10.98 & 36.86 & 19.00 & 9.16 & 88.05 \\
    MoCo-v1 + AugSelf~\cite{lee2021improving} & 64.94 & 37.01 & 32.84 & 33.13	 & 45.95 & 38.59 & 45.15 & 8.33 & 15.14 & 40.37 & 20.48 & 11.27 & 88.12   \\
    \textbf{MoCo-v1+\our{}} & \textbf{80.53} & \textbf{53.55} & \textbf{52.11} & \textbf{51.94} & \textbf{57.58} & \textbf{60.56} & \textbf{60.33} & \textbf{18.68} & \textbf{28.68} & \textbf{53.94} & \textbf{36.71} & \textbf{18.88} & \textbf{88.21}\\

    \bottomrule
    
    \end{tabular}
    }
\end{table*}

\end{document}